%% file: main.tex
\documentclass{article}

\usepackage{microtype}
\usepackage{graphicx}
\usepackage{subfigure}
\usepackage{booktabs} % for professional tables
\usepackage{amsfonts}
\usepackage{amsmath}
\usepackage{caption}
\usepackage{hyperref}
\usepackage{tabularx}

\usepackage{enumitem}

\usepackage[accepted]{icml2020}

\icmltitlerunning{ControlVAE: Controllable Variational Autoencoder}

\begin{document}

\twocolumn[
\icmltitle{ControlVAE: Controllable Variational Autoencoder}

% It is OKAY to include author information, even for blind
% submissions: the style file will automatically remove it for you
% unless you've provided the [accepted] option to the icml2020
% package.

% List of affiliations: The first argument should be a (short)
% identifier you will use later to specify author affiliations
% Academic affiliations should list Department, University, City, Region, Country
% Industry affiliations should list Company, City, Region, Country

% You can specify symbols, otherwise they are numbered in order.
% Ideally, you should not use this facility. Affiliations will be numbered
% in order of appearance and this is the preferred way.
% \icmlsetsymbol{equal}{*}

\begin{icmlauthorlist}
% \icmlauthor{Aeiau Zzzz}{equal,to}
% \icmlauthor{Bauiu C.~Yyyy}{equal,to,goo}
\icmlauthor{Huajie Shao}{uiuc}
\icmlauthor{Shuochao Yao}{uiuc}
\icmlauthor{Dachun Sun}{uiuc}
\icmlauthor{Aston Zhang}{am} \\
\icmlauthor{Shengzhong Liu }{uiuc}
\icmlauthor{Dongxin Liu }{uiuc}
\icmlauthor{Jun Wang}{ali}
\icmlauthor{Tarek Abdelzaher}{uiuc}

% \icmlauthor{Tateu H.~Yasehe}{ed,to,goo}
% \icmlauthor{Aaoeu Iasoh}{goo}
% \icmlauthor{Buiui Eueu}{ed}
% \icmlauthor{Aeuia Zzzz}{ed}
% \icmlauthor{Bieea C.~Yyyy}{to,goo}
% \icmlauthor{Teoau Xxxx}{ed} Shengzhong Liu 
% \icmlauthor{Eee Pppp}{ed} Dachun Sun
\end{icmlauthorlist}

\icmlaffiliation{uiuc}{Department of Computer Science, University of Illinois at Urbana-Champaign, Urbana, USA.}
\icmlaffiliation{am}{AWS Deep Learning, CA, USA.}
\icmlaffiliation{ali}{Alibaba Group at Seattle, Washington, USA}

\icmlcorrespondingauthor{Tarek Abdelzaher}{zaher@illinois.edu}
\icmlcorrespondingauthor{Huajie Shao}{hshao5@illinois.edu}

% You may provide any keywords that you
% find helpful for describing your paper; these are used to populate
% the "keywords" metadata in the PDF but will not be shown in the document
\icmlkeywords{Controllable VAE, PID, Generative Model, Control Theory}
\vskip 0.3in
]

% this must go after the closing bracket ] following \twocolumn[ ...

% This command actually creates the footnote in the first column
% listing the affiliations and the copyright notice.
% The command takes one argument, which is text to display at the start of the footnote.
% The \icmlEqualContribution command is standard text for equal contribution.
% Remove it (just {}) if you do not need this facility.

%\printAffiliationsAndNotice{}  % leave blank if no need to mention equal contribution
% \printAffiliationsAndNotice{\icmlEqualContribution} % otherwise use the standard text.

\printAffiliationsAndNotice
%\begin{abstract}
%This document provides a basic paper template and submission guidelines.
%Abstracts must be a single paragraph, ideally between 4--6 sentences long.
%Gross violations will trigger corrections at the camera-ready phase.
%\end{abstract}

%%%---set math formulat---
\setlength{\abovedisplayskip}{6.7pt}
\setlength{\belowdisplayskip}{6.7pt}
\setlength{\intextsep}{1\baselineskip}

\input{Abstract}
\input{introduction}

\input{preliminary}
\input{Model}
\input{Evaluation}

\input{relatedwork}

\input{Conclusion}
\input{acknowledge}

% Acknowledgements should only appear in the accepted version.
%\section*{Acknowledgements}
%
%\textbf{Do not} include acknowledgements in the initial version of
%the paper submitted for blind review.

% In the unusual situation where you want a paper to appear in the
% references without citing it in the main text, use \nocite
\nocite{langley00}

\bibliography{references}
\bibliographystyle{icml2020}

%%%%%%%%%%%%%%%%%%%%%%%%%%%%%%%%%%%%%%%%%%%%%%%%%%%%%%%%%%%%%%%%%%%%%%%%%%%%%%%
%%%%%%%%%%%%%%%%%%%%%%%%%%%%%%%%%%%%%%%%%%%%%%%%%%%%%%%%%%%%%%%%%%%%%%%%%%%%%%%
% DELETE THIS PART. DO NOT PLACE CONTENT AFTER THE REFERENCES!
%%%%%%%%%%%%%%%%%%%%%%%%%%%%%%%%%%%%%%%%%%%%%%%%%%%%%%%%%%%%%%%%%%%%%%%%%%%%%%%
%%%%%%%%%%%%%%%%%%%%%%%%%%%%%%%%%%%%%%%%%%%%%%%%%%%%%%%%%%%%%%%%%%%%%%%%%%%%%%%
\newpage
\input{appendix}

%\textbf{\emph{Do not put content after the references.}}
%%
%Put anything that you might normally include after the references in a separate
%supplementary file.
%
%We recommend that you build supplementary material in a separate document.
%If you must create one PDF and cut it up, please be careful to use a tool that
%doesn't alter the margins, and that doesn't aggressively rewrite the PDF file.
%pdftk usually works fine. 
%
%\textbf{Please do not use Apple's preview to cut off supplementary material.} In
%previous years it has altered margins, and created headaches at the camera-ready
%stage.
%%%%%%%%%%%%%%%%%%%%%%%%%%%%%%%%%%%%%%%%%%%%%%%%%%%%%%%%%%%%%%%%%%%%%%%%%%%%%%%
%%%%%%%%%%%%%%%%%%%%%%%%%%%%%%%%%%%%%%%%%%%%%%%%%%%%%%%%%%%%%%%%%%%%%%%%%%%%%%%

\end{document}

%% file: Abstract.tex
\begin{abstract}
Variational Autoencoders (VAE) and their variants have been widely used in a variety of applications, such as dialog generation, image generation and disentangled representation learning. However, the existing VAE models have some limitations in different applications. For example, a VAE easily suffers from KL vanishing in language modeling and low reconstruction quality for disentangling. To address these issues, we propose a novel controllable variational autoencoder framework, ControlVAE, that combines a controller, inspired by automatic control theory, with the basic VAE to improve the performance of resulting generative models. Specifically, we design a new non-linear PI controller, a variant of the proportional-integral-derivative (PID) control, to automatically tune the hyperparameter (weight) added in the VAE objective using the output KL-divergence as feedback during model training. The framework is evaluated using three applications; namely, language modeling, disentangled representation learning, and image generation. The results show that ControlVAE can achieve much better reconstruction quality than the competitive methods for the comparable disentanglement performance. For language modelling, it not only averts the KL-vanishing, but also improves the diversity of generated text. Finally, we also demonstrate that ControlVAE improves the reconstruction quality for image generation compared to the original VAE.
\end{abstract}

%% file: introduction.tex
\section{Introduction}
\label{sec:introduction}
This paper proposes a novel controllable variational autoencoder, ControlVAE~\footnote{Source code is publicly available at~\url{https://github.com/shj1987/ControlVAE-ICML2020.git}}, that leverages automatic control to precisely control the trade-off between data reconstruction accuracy bounds (from a learned latent representation) and application-specific constraints, such as output diversity or disentangled latent factor representation. Specifically, a controller is designed that stabilizes the value of KL-divergence (between the learned approximate distribution of the latent variables and their true distribution) in the VAE's objective function to achieve the desired trade-off, thereby improving application-specific performance metrics of several existing 
VAE models.

The work is motivated by the increasing popularity of VAEs as an unsupervised generative modeling framework that learns an approximate mapping between Gaussian latent variables and data samples when the true latent variables have an intractable posterior distribution ~\cite{sohn2015learning,kingma2013auto}. Since VAEs can directly work with both continuous and discrete input data~\cite{kingma2013auto}, they have been widely adopted in various applications, such as image generation~\cite{yan2016attribute2image,liu2017unsupervised}, dialog generation~\cite{wang2019topic,hu2017toward}, and disentangled representation learning~\cite{higgins2017beta,kim2018disentangling}.

Popular VAE applications often involve a trade-off between reconstruction accuracy bounds and some other application-specific goal, effectively manipulated through KL-divergence. For example, in (synthetic) text or image generation, a goal is to produce {\em new original text or images\/}, as opposed to reproducing one of the samples in training data. In text generation, if KL-divergence is too low, output diversity is reduced~\cite{bowman2015generating}, which is known as the KL-vanishing problem. To increase output diversity, it becomes advantageous to artificially {\em increase KL-divergence\/}. The resulting approximation was shown to produce more diverse, yet still authentic-looking outputs. Conversely, disentangled representation learning~\cite{denton2017unsupervised} leverages the observation that KL-divergence in the VAE constitutes an upper bound on information transfer through the latent channels per data sample~\cite{burgess2018understanding}. Artificially {\em decreasing KL-divergence\/} (e.g., by increasing its weight in a VAE's objective function, which is known as the $\beta$-VAE) therefore imposes a stricter information bottleneck, which was shown to force the learned latent factors to become more independent (i.e., non-redundant), leading to a better disentangling. 
The above examples suggest that a useful extension of VAEs is one that allows users to exercise explicit control over KL-divergence in the objective function. ControlVAE realizes this extension.

We apply ControlVAE to three different applications: language modeling, disentangling, and image generation. Evaluation results on real-world datasets demonstrate that ControlVAE is able to achieve an {\em adjustable trade-off\/} between reconstruction error and KL-divergence. It can discover more disentangled factors and significantly reduce the reconstruction error compared to the $\beta$-VAE~\cite{burgess2018understanding} for disentangling. For language modeling, it can not only completely avert the KL vanishing problem, but also improve the diversity of generated data. Finally, we also show that ControlVAE improves the reconstruction quality on image generation task via slightly increasing the value of KL divergence compared with the original VAE.

%% file: preliminary.tex
\section{Preliminaries}
\label{sec:preliminary}
The objective function of VAEs consists of two terms: log-likelihood and KL-divergence. The first term tries to reconstruct the input data, while KL-divergence has the desirable effect of keeping the representation of input data sufficiently diverse. In particular, KL-divergence can affect both the reconstruction quality and diversity of generated data. If the KL-divergence is too high, it would affect the accuracy of generated samples. If it is too low, output diversity is reduced, which may be a problem in some applications such as language modeling~\cite{bowman2015generating} (where it is known as the KL-vanishing problem). 

To mitigate KL vanishing, one promising way is to add an extra hyperparameter $\beta (0 \leq \beta \leq 1)$ in the VAE objective function to control the KL-divergence via increasing $\beta$ from $0$ until to $1$ with sigmoid function or cyclic function~\cite{liu2019cyclical}. These methods, however, blindly change $\beta$ without sampling the actual KL-divergence during model training. Using a similar methodology, researchers recently developed a new $\beta$-VAE ($\beta > 1$)~\cite{higgins2017beta,burgess2018understanding} to learn the disentangled representations by controlling the value of KL-divergence. However, $\beta$-VAE suffers from high reconstruction errors~\cite{kim2018disentangling}, because it adds a very large $\beta$ in the VAE objective so the model tends to focus disproportionately on optimizing the KL term. In addition, its hyperparameter is fixed during model training, missing the chance of balancing the reconstruction error and KL-divergence.

\begin{figure}[!tb]
%\vskip 0.05in
\begin{center}
\centerline{\includegraphics[width=0.98\columnwidth]{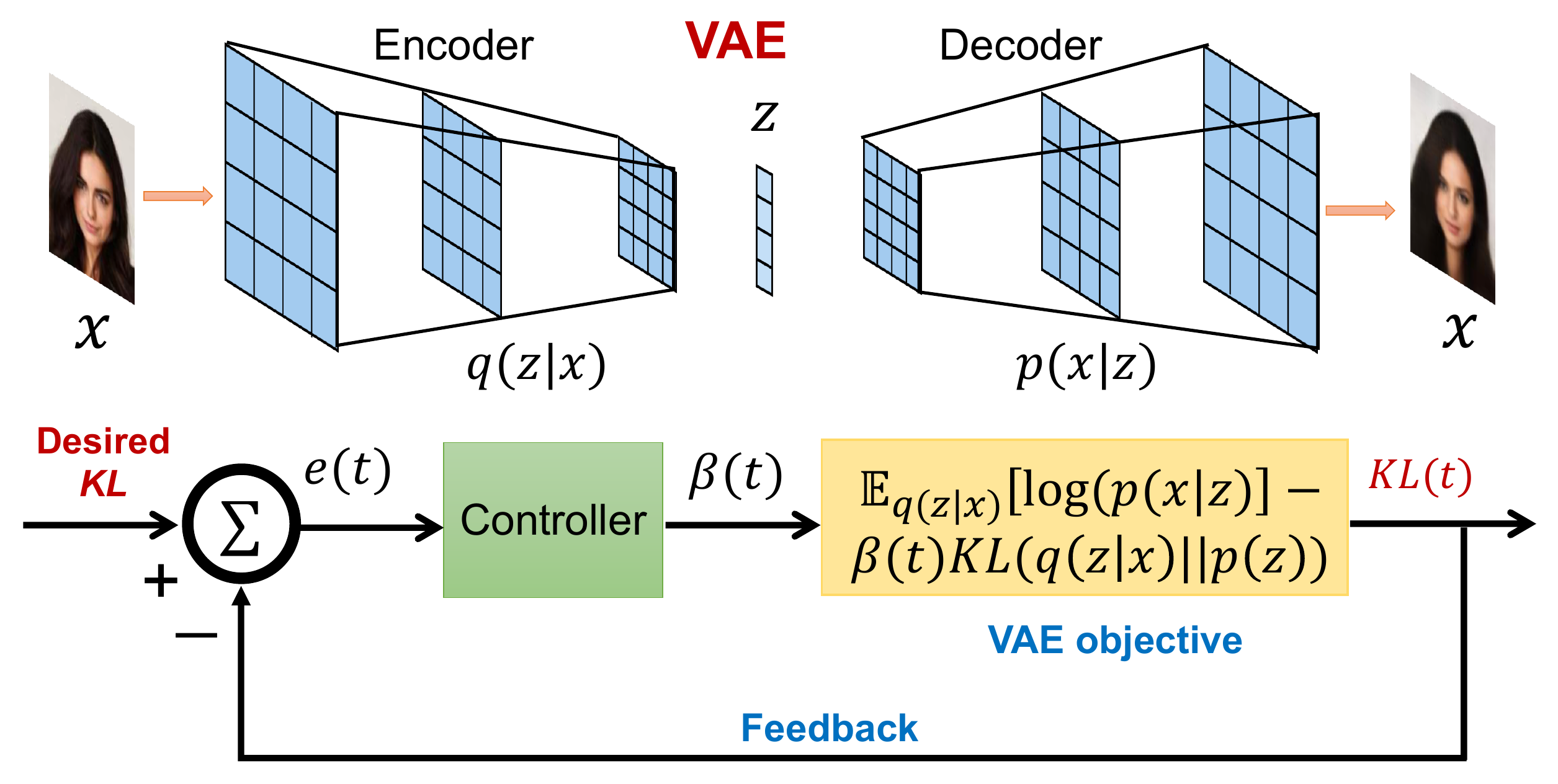}}
\vskip -0.05in
\caption{Framework of ControlVAE. It combines a controller with the basic VAE framework to stabilize the KL divergence to a specified value via automatically tuning the weight $\beta(t)$ in the objective.}
\label{fig:controlVAE}
\end{center}
\vskip -0.35in
\end{figure}

The core technical challenge responsible for the above application problems lies in the difficulty to tune the weight of the KL-divergence term during model training. Inspired by control systems, we fix this problem using feedback control. Our controllable variational autoencoder is illustrated in Fig.~\ref{fig:controlVAE}. It samples the output KL-divergence at each training step $t$, and feeds it into an algorithm that tunes the hyperparameter, $\beta(t)$, accordingly, aiming to stabilize KL-divergence at a desired value, called the \textit{set point}.

We further design a non-linear PI controller, a variant of the PID control algorithm~\cite{aastrom2006advanced}, to tune the hyperparameter $\beta(t)$. PID control is the basic and most prevalent form of feedback control in a large variety of industrial~\cite{aastrom2006advanced} and software performance control~\cite{hellerstein2004feedback} applications. The general model of PID controller is defined by
\begin{equation}\label{eq:pid}
\beta(t) = K_p e(t) + K_i \int_0^t e(\tau)d\tau + K_d \frac{de(t)}{dt},
\end{equation}
\noindent
where $\beta(t)$ is the output of the controller; $e(t)$ is the error between the actual value and the desired value at time $t$; $K_p, K_i$ and $K_d$ denote the coefficients for the P term, I term and D term, respectively.

The basic idea of the PID algorithm is to calculate an error, $e(t)$, between a set point (in this case, the desired KL-divergence) and the current value of the controlled variable (in this case, the actual KL-divergence), then apply a correction in a direction that reduces that error. The correction is applied to some intermediate directly accessible variable (in our case, $\beta(t)$) that influences the value of the variable we ultimately want to control (KL-divergence). In general, the correction computed by the controller is the weighted sum of three terms; one changes with error (P), one changes with the integral of error (I), and one changes with the derivative of error (D). In a nonlinear controller, the changes can be described by {\em nonlinear\/} functions. Note that, since derivatives essentially compute the slope of a signal, when the signal is noisy, the slope often responds more to variations induced by noise. Hence, following established best practices in control of noisy systems, we do not use the derivative (D) term in our specific controller. 
Next, we introduce VAEs and our objective in more detail.
%We introduce a strategy to tune our non-linear controller to ensure the stability of model training. We also provide theoretical upper and lower bounds to inform the choice of set point on KL-divergence that ControlVAE should maintain. Let us now begin with the problem statement.

\subsection{The Variational Autoencoder (VAE)}
\label{sec:vae-intro}
Suppose that we have a dataset $\mathbf{x}$ of $n$ i.i.d. samples that are generated by the ground-truth latent variable $\mathbf{z}$, interpreted as the representation of the data. Let $p_\theta (\mathbf{x}|\mathbf{z})$ denote a probabilistic \textit{decoder} with a neural network to generate data $\mathbf{x}$ given the latent variable $\mathbf{z}$. The distribution of representation corresponding to the dataset $\mathbf{x}$ is approximated by the variational posterior, $q_{\phi} (\mathbf{z}|\mathbf{x})$, which is produced by an \textit{encoder} with a neural network. 
The Variational Autoencoder (VAE)~\cite{rezende2014stochastic, kingma2013auto} has been one of the most popular generative models. The basic idea of VAE can be summarized in the following: (1) VAE encodes the input data samples $\mathbf{x}$ into a latent variable $\mathbf{z}$ as its distribution of representation via a probabilistic encoder, which is parameterised by a neural network. (2) then adopts the decoder to reconstruct the original input data based on the samples from $\mathbf{z}$. VAE tries to maximize the marginal likelihood of the reconstructed data, but it involves intractable posterior inference. Thus, researchers adopt backpropagation and stochastic gradient descent~\cite{kingma2013auto} to optimize its variational lower bound of log likelihood~\cite{kingma2013auto}.
\begin{equation}\label{eq:vae}
\begin{split}
& \log p_\theta (\mathbf{x})  \geq \mathcal{L}_{vae} \\
& =  \mathbb{E}_{q_\phi(\mathbf{z|x)}} [\log p_\theta(\mathbf{x|z})] - D_{KL} (q_\phi(\mathbf{z|x})||p(\mathbf{z})),
\end{split}
\end{equation}
where $p(\mathbf{z})$ is the prior distribution, e.g., Unit Gaussian. $p_\theta (\mathbf{x}|\mathbf{z})$ is a probabilistic \textit{decoder} parameterized by a neural network to generate data $\mathbf{x}$ given the latent variable $\mathbf{z}$, and the posterior distribution of latent variable $\mathbf{z}$ given data $\mathbf{x}$ is approximated by the variational posterior, $q_{\phi} (\mathbf{z}|\mathbf{x})$, which is parameterized by an \textit{encoder} network. The VAE is trained by maximizing $\mathcal{L}_{vae}$, which consists of a reconstruction term and a KL term, over the training data.

However, the basic VAE models cannot explicitly control the KL-divergence to a specified value. They also often suffer from KL vanishing (in language modeling~\cite{bowman2015generating,liu2019cyclical}), which means the KL-divergence becomes zero during optimization. 
%
%To remedy this issue, one popular way is to add a hyperparameter $\beta$ on the KL term~\cite{bowman2015generating,liu2019cyclical}, and then gradually increases it from $0$ until $1$. However, the existing methods, such as KL cost annealing and cyclical annealing~\cite{bowman2015generating,liu2019cyclical}, cannot totally avert KL vanishing because they blindly vary the hyperparameter $\beta$ during model training.

\subsection{$\beta$-VAE}
\label{sec:beta-vae}
$\beta$-VAE~\cite{higgins2017beta,chen2018isolating} is an extension to the basic VAE framework, often used as an unsupervised method for learning a disentangled representation of the data generative factors. A disentangled representation, according to the literature~\cite{bengio2013representation}, is defined as one where single latent units are sensitive to changes in single generative factors, while being relatively invariant to changes in other factors.
Compared to the original VAE, $\beta$-VAE adds an extra hyperparameter $\beta (\beta >1 )$ as a weight of KL-divergence in the original VAE objective~\eqref{eq:vae}. It can be expressed by
\begin{equation}\label{eq:beta-vae}
\mathcal{L}_{\beta} = \mathbb{E}_{q_\phi(\mathbf{z|x)}} [\log p_\theta(\mathbf{x|z})] - \beta D_{KL} (q_\phi(\mathbf{z|x})||p(\mathbf{z})).
\end{equation}
In order to discover more disentangled factors, researchers further put a constraint on total information capacity, $C$, to control the capacity of the information bottleneck (KL-divergence)~\cite{burgess2018understanding}. Then Lagrangian method is adopted to solve the following optimization problem.
\begin{equation}\label{eq:improved-vae}\small
\mathcal{L}_{\beta} = \mathbb{E}_{q_\phi(\mathbf{z|x)}} [\log p_\theta(\mathbf{x|z})] - \beta | D_{KL} (q_\phi(\mathbf{z|x})||p(\mathbf{z})) -C |,
\end{equation}
where $\beta$ is a large hyperparameter (e.g., 100).

However, one drawback of $\beta$-VAE is that it obtains good disentangling at the cost of reconstruction quality. When the weight $\beta$ is large, the optimization algorithm tends to optimize the second term in \eqref{eq:improved-vae}, leading to a high reconstruction error.

The above background suggests that a common challenge in applying VAEs (and their extensions) lies in appropriate weight allocation among the reconstruction accuracy and KL-divergence in the VAEs objective function. As mentioned earlier, we solve this using a nonlinear PI controller that manipulates the value of the non-negative hyperparameter, $\beta(t)$. This algorithm is described next.

%% file: Model.tex
\section{The ControlVAE Algorithm}
\label{sec:model}
During model training, we sample the output KL-divergence, which we denote by $\hat{v}_{kl}(t)$, at training step $t$. The sampled KL-divergence is then compared to the set point, ${v}_{kl}$, and the difference, $e(t) = {v}_{kl} - \hat{v}_{kl}(t)$ then used as the feedback to a controller to calculate the hyperparameter $\beta(t)$. ControlVAE can be expressed by the following variational lower bound:
\begin{equation}\label{eq:controlVAE-object}
\mathcal{L} = \mathbb{E}_{q_\phi(\mathbf{z}|\mathbf{x})} [\log p_\theta(\mathbf{x|z})] - \beta(t) D_{KL} (q_\phi(\mathbf{z}|\mathbf{x})||p(\mathbf{z})),
\end{equation}

%%%%--design PI control algorithm
When KL-divergence drops below the set point, the controller counteracts this change by reducing the hyperparameter $\beta(t)$ (to reduce penalty for KL-divergence in the objective function~\eqref{eq:controlVAE-object}). The reduced weight, $\beta(t)$, allows KL-divergence to grow, thus approaching the set point again. Conversely, when KL-divergence grows above the set point, the controller increases $\beta(t)$ (up to a certain value), thereby increasing the penalty for KL-divergence and forcing it to decrease. This effect is achieved by computing $\beta(t)$ using Equation~(\ref{eq:vae-lang}), below, which is an instance of nonlinear PI control: \begin{equation}\label{eq:vae-lang}
\vspace{-0.02in}
\beta(t) = \frac{K_p}{1+\exp(e(t))} - K_i \sum_{j=0}^t e(j) + \beta_{min},
\end{equation}
where $K_p$ and $K_i$ are the constants. The first term (on the right hand side) ranges between $0$ and $K_p$ thanks to the exponential function $exp(.)$. Note that when error is large and positive (KL-diverge is below set point), the first term approaches 0, leading to a lower $\beta(t)$ that encourages KL-divergence to grow. Conversely, when error is large and negative (KL-divergence above set point), the first term approaches its maximum (which is $K_p$), leading to a higher $\beta(t)$ that encourages KL-divergence to shrink.

\begin{figure}[!t]
\begin{center}
\centerline{\includegraphics[width=0.99\columnwidth]{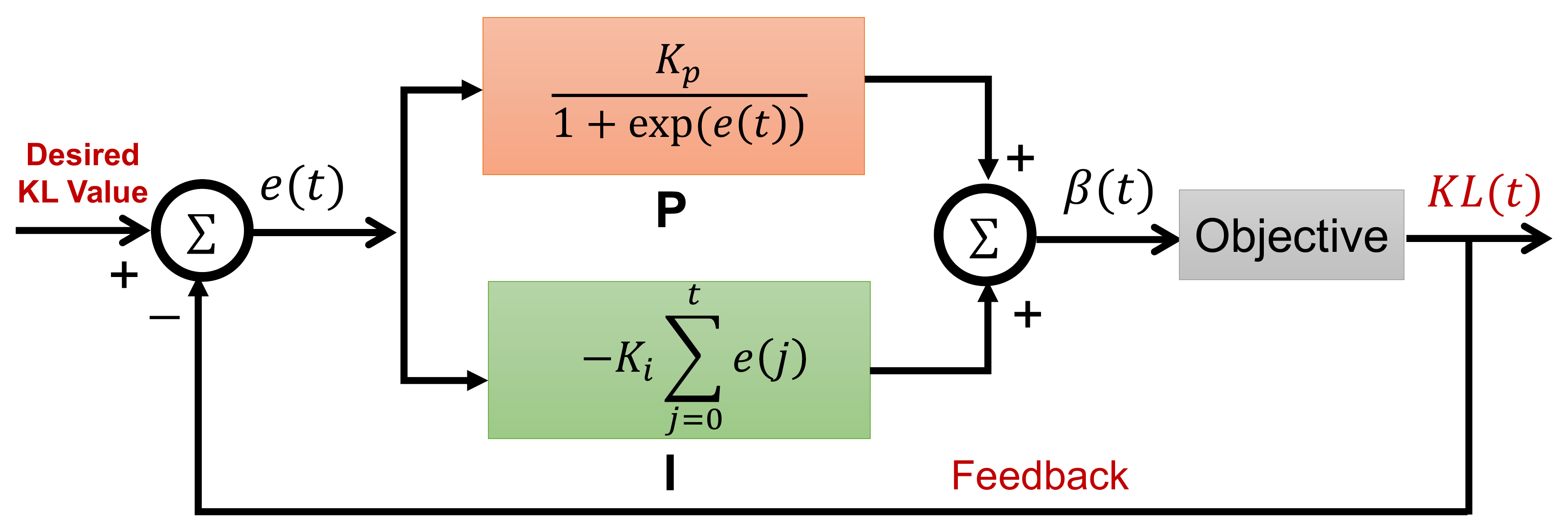}}
\vskip -0.05in
\caption{PI controller. It uses the output KL-divergence at training step $t$ as the feedback to the PI algorithm to compute $\beta(t)$.}
\label{fig:pid}
\end{center}
\vskip -0.25in
\end{figure}

The second term of the controller sums (integrates) past errors with a sampling period $T$ (one training step in this paper). This creates a progressively stronger correction (until the sign of the error changes). The negative sign ensures that while errors remain positive (i.e., when KL-divergence is below set point), this term continues to decrease, whereas while errors remain negative (i.e., when KL-divergence is above set point), this term continues to increase. In both cases, the change forces $\beta(t)$ in a direction that helps KL-divergence approach the set point. In particular, note that when the error becomes zero, the second term (and thus the entire right hand side) stops changing, allowing controller output, $\beta(t)$, to stay at the same value that hopefully caused the zero error in the first place. This allows the controller to ``lock in" the value of $\beta(t)$ that meets the KL-divergence set point.
Finally, $\beta_{min}$ is an application-specific constant. It effectively shifts the range within which $\beta(t)$ is allowed to vary. This PI controller is illustrated in Fig.~\ref{fig:pid}.

%%figure%%

%%%%% PI tuning %%%%

\subsection{PI Parameter Tuning for ControlVAE}
\label{sec:pi_tuning}
One challenge of applying the PI control algorithm lies how to tune its parameters, $K_p$ and $K_i$ effectively. While optimal tuning of nonlinear controllers is non-trivial, in this paper we follow a very simple rule: tune these constants to ensure that reactions to errors are sufficiently smooth to allow gradual convergence.
Let us first consider the coefficient $K_p$. Observe that the maximum (positive) error occurs when actual KL-divergence is close to zero. In this case, if $v_{kl}$ is the set point on KL-divergence, then the error, $e(t)$, is approximated by $e(t) \approx v_{kl} - 0 = v_{kl}$. When KL-divergence is too small, the VAE does not learn useful information from input data~\cite{liu2019cyclical}. We need to assign $\beta(t)$ a very small non-negative value, so that KL-divergence is encouraged to grow (when the resulting objective function is optimized). In other words, temporarily ignoring other terms in Equation~\eqref{eq:vae-lang}, the contribution of the first term alone should be sufficiently small:
\begin{equation}\label{eq:Kp}
\frac{K_p}{1+\exp( v_{kl} )} \leq \epsilon,
\end{equation}
where $\epsilon$ is a small constant (e.g., $10^{-3}$ in our implementation). The above~\eqref{eq:Kp} can also be rewritten as $K_p \leq (1+\exp(v_{kl}))\epsilon$. Empirically, we find that $K_p=0.01$ leads to good performance and satisfies the above constraint.

Conversely, when the actual KL-divergence is much larger than the desired value $v_{kl}$, the error $e(t)$ becomes a large negative value. As a result, the first term in~\eqref{eq:vae-lang} becomes close to a constant, $K_p$. If the resulting larger value of $\beta(t)$ is not enough to cause KL-divergence to shrink, one needs to gradually continue to increase $\beta(t)$. This is the job of second term. 
The negative sign in front of that term ensures that when negative errors continue to accumulate, the positive output $\beta(t)$ continues to increase. Since it takes lots of steps to train deep VAE models,
the increase per step should be very small, favoring smaller values of $K_i$. 
Empirically we found that a value $K_i$ between $10^{-3}$ and $10^{-4}$ stabilizes the training. Note that, $K_i$ should not be too small either, because it would then unnecessarily slow down the convergence.

%%%%%bound of set point %%%
\subsection{Set Point Guidelines for ControlVAE}
\label{sec:bound}
The choice of desired value of KL-divergence (set point) is largely application specific. In general, when $ \beta_{min} \leq \beta(t) \leq \beta_{max}$, the upper bound of expected KL-divergence is the value of KL-divergence as ControlVAE converges when $\beta(t) = \beta_{min}$, denoted by $V_{max}$. Similarly, its lower bound, $V_{min}$, can be defined as the KL-divergence produced by ControlVAE when $\beta(t) = \beta_{max}$. For feedback control to be most effective (i.e., not run against the above limits), the KL-divergence set point should vary in the range of $[V_{min}, V_{max}]$. Since ControlVAE is an end-to-end learning model, users can customize the desired value of KL-divergence to meet their demand with respect to different applications. For instance, if some users prefer to improve the diversity of text generation and image generation, they can slightly increase the KL-divergence. Otherwise they can reduce the KL-divergence if they want to improve the generation accuracy.

%%%%%algorithm%%%
\subsection{Summary of the PI Control Algorithm}
We summarize the proposed PI control algorithm in Algorithm~\ref{alg:pid}. Our PI algorithm updates the hyperparameter, $\beta(t)$, with the feedback from sampled KL-divergence at training step $t$. Line $6$ computes the error between the desired KL-divergence, $v_{kl}(t)$, and the sampled $\hat{v}_{kl}(t)$. Line $7$ to $9$ calculate the P term and I term for the PI algorithm, respectively. Note that, Line 10 and 11 is a popular constraint in PID/PI design, called anti-windup~\citep{azar2015design,peng1996anti}. It effectively disables the integral term of the controller when controller output gets out of range, not to exacerbate the out-of-range deviation. Line $13$ is the calculated hyperparameter $\beta(t)$ by PI algorithm in~\eqref{eq:vae-lang}. Finally, Line $14$ to $19$ aim to limit $\beta(t)$ to a certain range, $[\beta_{min},\beta_{max}]$.

%%%--PID algorithm----
% \vspace{-0.1in}
\begin{algorithm}[!tb]
   \caption{PI algorithm.}
   \label{alg:pid}
\begin{algorithmic}[1]\small
   \STATE {\bfseries Input:} desired KL $v_{kl}$, coefficients $K_p$, $K_i$, max/min value $\beta_{max}$, $\beta_{min}$, iterations $N$
   \STATE {\bfseries Output:} hyperparameter $\beta(t)$ at training step $t$
%   %\REPEAT
   \STATE {\bfseries Initialization}: $I(0)=0$, $\beta(0)=0$
   \FOR{$t=1$ {\bfseries to} $N$}
   \STATE Sample KL-divergence, $\hat{v}_{kl}(t)$
   \STATE $e(t) \leftarrow v_{kl}-\hat{v}_{kl}(t)$
   \STATE $P(t) \leftarrow \frac{K_p}{1+\exp(e(t))}$
   \IF{ $ \beta_{min} \leq \beta(t-1) \leq \beta_{max}$ }
   		\STATE $I(t) \leftarrow I(t-1) - K_i e(t)$
   \ELSE
   \STATE $I(t) = I(t-1)$ \quad // Anti-windup
	 \ENDIF
   \STATE $\beta(t)=P(t)+I(t) + \beta_{min}$
   \IF{$\beta(t)> \beta_{max}$}
   \STATE $\beta(t)= \beta_{max}$
 	\ENDIF
   \IF{$\beta(t)< \beta_{min}$}
   \STATE $\beta(t)= \beta_{min}$
   \ENDIF
   \STATE \textbf{Return} $\beta(t)$
 \ENDFOR
\end{algorithmic}
\end{algorithm}
\vspace{-0.1in}

\subsection{Applications of ControlVAE}
As a preliminary demonstration of the general applicability of the above approach and as an illustration of its customizability, we apply ControlVAE to three different applications stated below.
% : language modeling, disentangling and image generation.
\begin{itemize}
\vspace{-0.1in}
%%%%%--language modeling
\item \textbf{Language modeling}: We first apply ControlVAE to solve the KL vanishing problem meanwhile improve the diversity of generated data. As mentioned in Section~\ref{sec:vae-intro}, the VAE models often suffer from KL vanishing in language modeling. The existing methods cannot completely solve the KL vanishing problem or explicitly manipulate the value of KL-divergence. In this paper, we adopt ControlVAE to control KL-divergence to a specified value to avoid KL vanishing using the output KL-divergence. Following PI tuning strategy in Section~\ref{sec:pi_tuning}, we set $K_p$, $K_i$ of the PI algorithm in~\eqref{eq:vae-lang} to $0.01$ and $0.0001$, respectively. In addition, $\beta_{min}$ is set to $0$ and the maximum value of $\beta(t)$ is limited to $1$.

%%%% disentangling%%%%%%
\item \textbf{Disentangling}: We then apply the ControlVAE model to achieve a better trade-off between reconstruction quality and disentangling. As mentioned in Section~\ref{sec:beta-vae}, $\beta$-VAE ($\beta>1$) assigns a large hyperparameter to the objective function to control the KL-divergence (information bottleneck), which, however, leads to a large reconstruction error. To mitigate this issue, we adopt ControlVAE to automatically adjust the hyperparameter $\beta(t)$ based on the output KL-divergence during model training. Using the similar methodology in~\cite{burgess2018understanding}, we train a single model by gradually increasing KL-divergence from $0.5$ to a desired value $C$ with a step function $\alpha$ for every $K$ training steps. Since $\beta(t)>1$, we set $\beta_{min}$ to $1$ for the PI algorithm in~\eqref{eq:vae-lang}. Following the PI tuning method above, the coefficients $K_p$ and $K_i$ are set to $0.01$ and $0.001$, respectively.
%%%%%%%%image generation
\item \textbf{Image generation}: The basic VAE models tend to produce blurry and unrealistic samples for image generation~\cite{zhao2017towards}. In this paper, we try to leverage ControlVAE to manipulate (slightly increase) the value of KL-divergence to improve the reconstruction quality for image generation. Different from the original VAE ($\beta(t)=1$), we extend the range of the hyperparameter, $\beta(t)$, from $0$ to $1$ in our controlVAE model. Given a desired KL-divergence, controlVAE can automatically tune $\beta(t)$ within that range. For this task, we use the same PI control algorithm and hyperparameters as the above language modeling.
\end{itemize}

%% file: Evaluation.tex
\section{Experiments}
\label{sec:evaluate}
We evaluate the performance of ControlVAE on benchmark datasets in the three different applications mentioned above. 
%Source code is publicly available at~\url{https://github.com/shj1987/ControlVAE-ICML2020.git}

%%%--figure for KL vanishing-
\begin{figure*}[!t]
\centering     %%% not \center
\subfigure[KL divergence]{\includegraphics[width=0.33\textwidth]{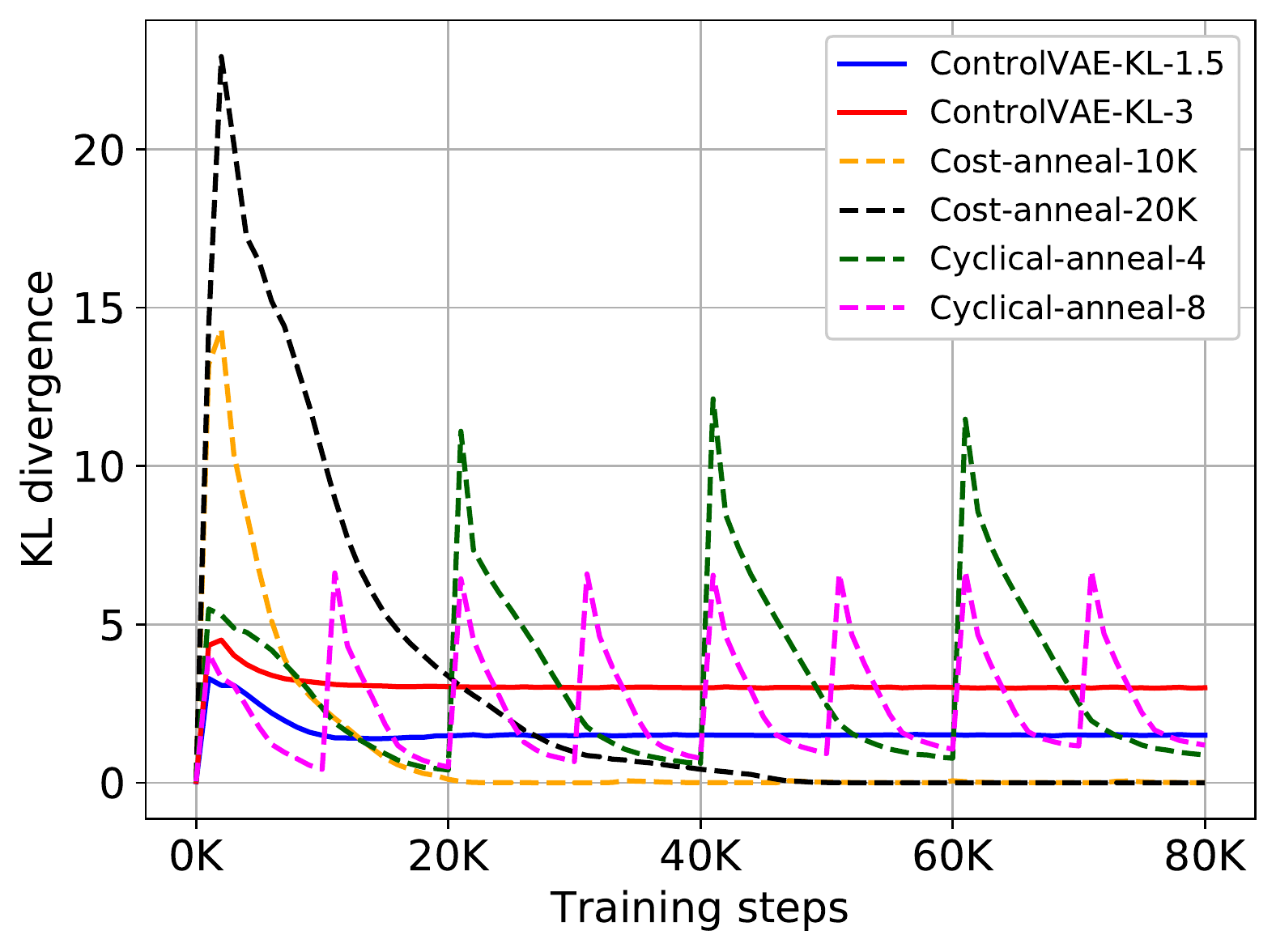}}
\subfigure[Reconstruction loss]{\includegraphics[width=0.33\textwidth]{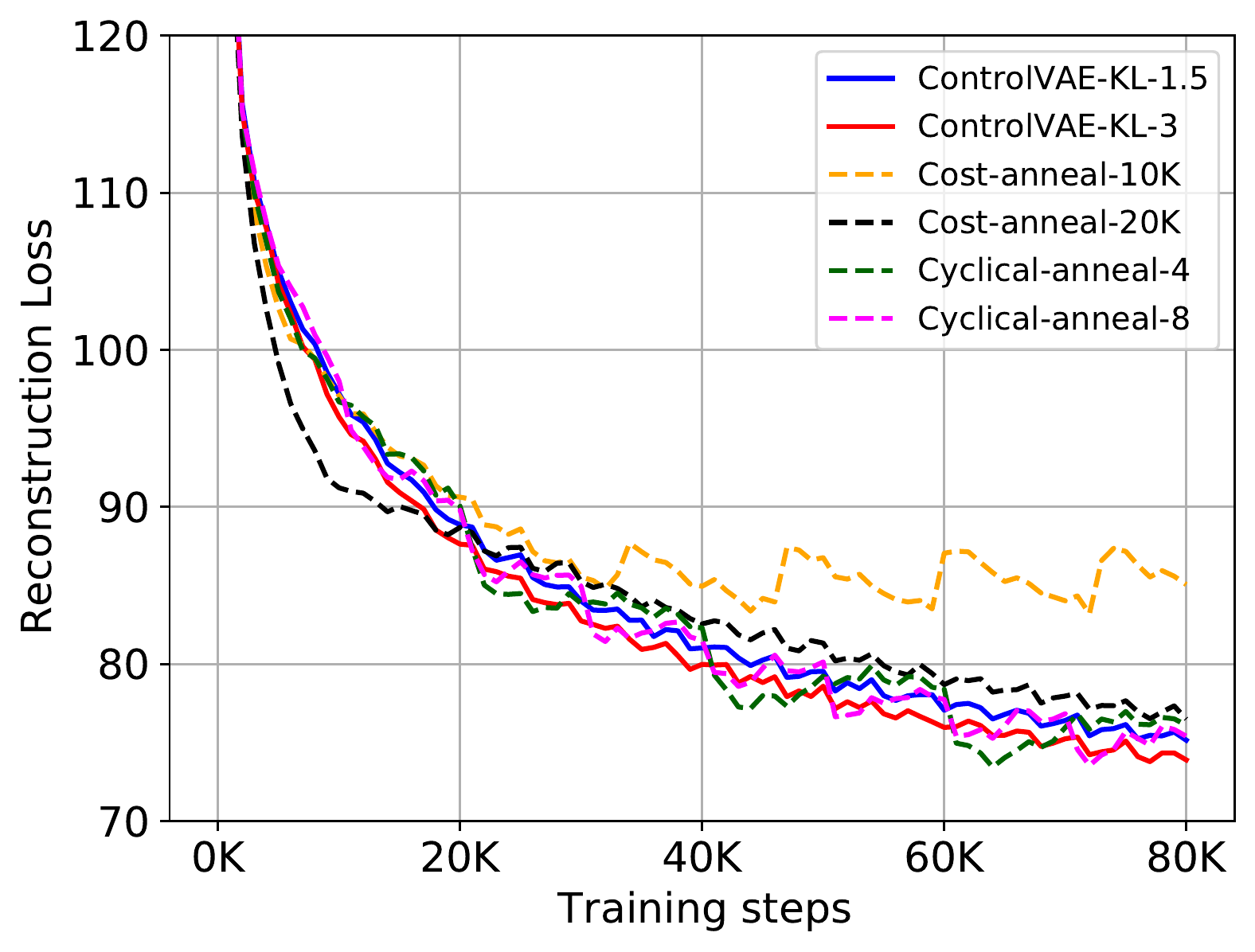}}
\subfigure[$\beta(t)$]{\label{fig:c}\includegraphics[width=0.33\textwidth]{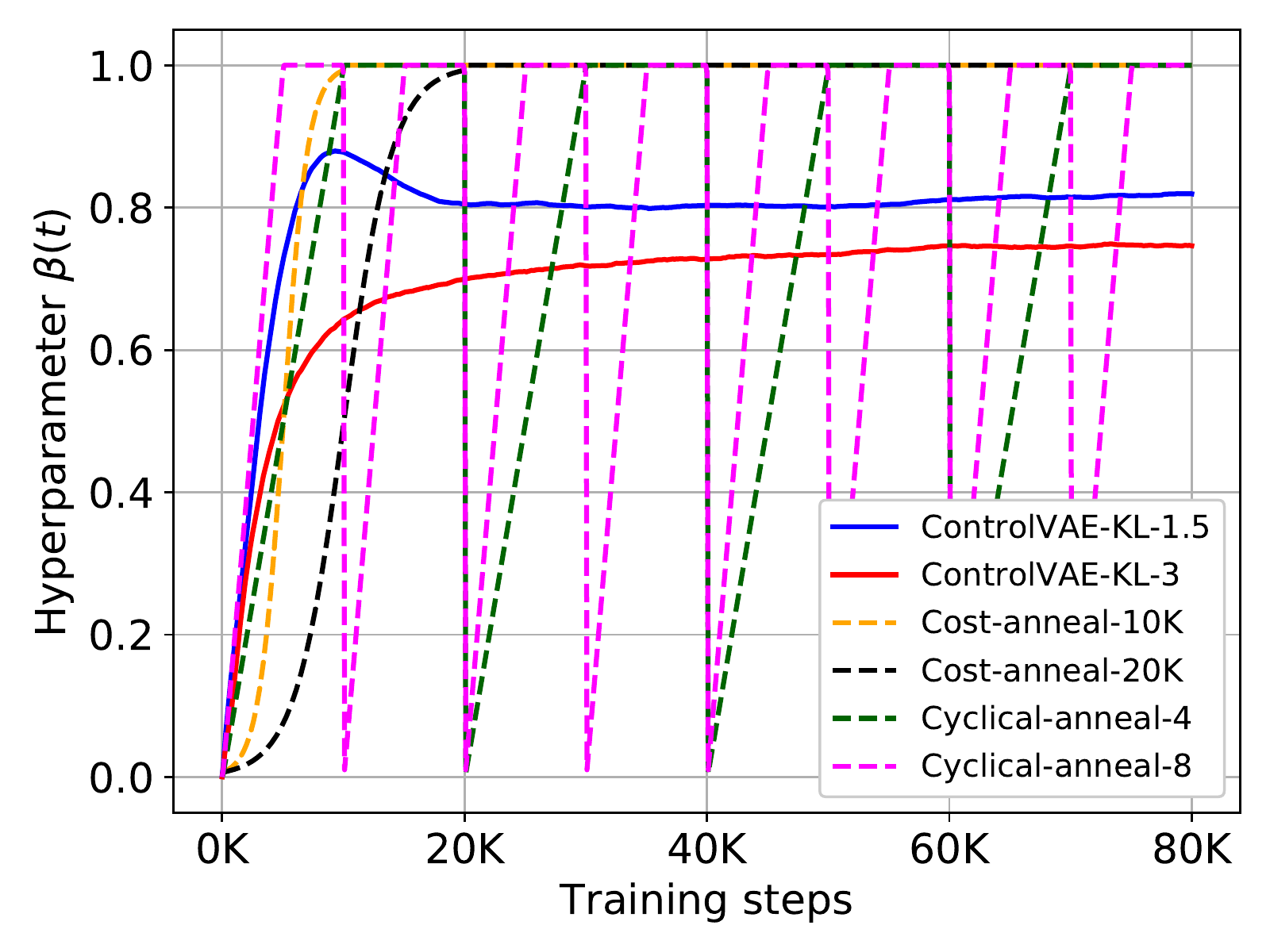}}
\vskip -0.1in
\caption{Performance comparison for different methods on the PTB data. (a) shows that ControlVAE and Cyclical annealing ($4, 8$ cycles) can avert KL vanishing, while Cost annealing still suffers from KL vanishing after $20K$ and $50K$ training steps. Moreover, ControlVAE can control the KL-divergence and also has lower reconstruction errors than the other methods in (b).}\label{fig:KL-vanish}
\vskip -0.05in
\end{figure*}
%%--end of figure

%%----table--%%
\begin{table*}[!tb]
\caption{Performance comparison for different methods on dialog-generation using SW data over $5$ random seeds. Dis-$n$: higher is better. PPL: lower is better, and self-BLEU lower is better.}
\label{tab:dist}
%\vskip 0.15in
\begin{center}
\begin{small}
% \begin{sc}
\begin{tabular}{llllll}
\toprule
Methods/metric & Dis-1 & Dis-2 & self-BLEU-2 & self-BLEU-3 & PPL \\
\midrule
ControlVAE-KL-35  & \textbf{6.27K} $\pm$ 41 &  \textbf{95.86K} $\pm$ 1.02K & \textbf{0.663} $\pm$  0.012 & \textbf{0.447} $\pm$ 0.013 & \textbf{8.81} $\pm$ 0.05  \\
ControlVAE-KL-25 & 6.10K $\pm$ 60 & 83.15K $\pm$ 4.00K & 0.698 $\pm$ 0.006 & 0.495 $\pm$  0.014  &  12.47 $\pm$ 0.07 \\
Cost anneal-KL-17  & 5.71K $\pm$ 87 & 69.60K $\pm$ 1.53K & 0.721 $\pm$ 0.010 & 0.536 $\pm$ 0.008 & 16.82 $\pm$ 0.11 \\
Cyclical (KL = 21.5)  & 5.79K $\pm$ 81 & 71.63K $\pm$ 2.04K &  0.710 $\pm$ 0.007 & 0.524 $\pm$  0.008 & 17.81 $\pm$ 0.33 \\
\bottomrule
\end{tabular}
% \end{sc}
\end{small}
\end{center}
\vskip -0.1in
\end{table*}
%%--end of table--%%

\subsection{Datasets}
The datasets used for our experiments are introduced below.
\begin{itemize}[noitemsep]
\vspace{-0.1in}
\item Language modeling: 1) \textbf{Penn Tree Bank (PTB)}~\cite{marcus1993building}: it consists of $42,068$ training sentences, $3,370$ validation sentences and $3,761$ testing sentences. 2) \textbf{Switchboard(SW)}~\cite{godfrey1997switchboard}: it has $2400$ two-sided telephone conversations with manually transcribed speech and alignment. The data is randomly split into $2316$, $60$ and $62$ dialog for training, validation and testing.
\item Disentangling: 1)~\textbf{2D Shapes}~\cite{matthey2017dsprites}: it has $737,280$ binary $64 \times 64$ images of 2D shapes with five ground truth factors (number of values): shape(3), scale(6), orientation(40), x-position(32), y-position(32)~\cite{kim2018disentangling}.
% 2)~\textbf{3D Chairs}~\cite{aubry2014seeing}: it consists of $86,366$ RGB $64 \times 64 \times 3$ images of chair CAD.
\item Image generation: 1) \textbf{CelebA}(cropped version)~\cite{liu2015deep}: It has $202,599$ RGB $128 \times 128 \times 3$ images of celebrity faces. The data is split into $192,599$ and $10,000$ images for training and testing.
\end{itemize}

\subsection{Model Configurations}
The detailed model configurations and hyperparameter settings for each model is presented in Appendix~\ref{sec:configure}.

%%--language model--
\subsection{Evaluation on Language Modeling}
First, we compare the performance of ControlVAE with the following baselines for mitigating KL vanishing in text generation~\cite{bowman2015generating}. 
\hfill \break
\textbf{Cost annealing}~\cite{bowman2015generating}: This method gradually increases the hyperparameter on KL-divergence from $0$ until to $1$ after $N$ training steps using sigmoid function.\hfill \break
\textbf{Cyclical annealing}~\cite{liu2019cyclical}: This method splits the training process into $M$ cycles and each increases the hyperparameter from $0$ until to $1$ using a linear function.

Fig.~\ref{fig:KL-vanish} illustrates the comparison results of KL-divergence, reconstruction loss and hyperparamter $\beta(t)$ for different methods on the PTB dataset. Note that, here ControlVAE-KL-$v$ means we set the KL-divergence to a desired value $v$ (e.g., 3) for our PI controller following the set point guidelines in Section~\ref{sec:bound}. Cost-annealing-$v$ means we gradually increase the hyperparameter, $\beta(t)$, from $0$ until to $1$ after $v$ steps using sigmoid function. We observe from Fig.~\ref{fig:KL-vanish}(a) that ControlVAE (KL=1.5, 3) and Cyclical annealing ($4, 8$ cycles) can avert the KL vanishing. However, our ControlVAE is able to stabilize the KL-divergence while cyclical annealing could not. Moreover, our method has a lower reconstruction loss than the cyclical annealing in Fig.~\ref{fig:KL-vanish} (b). Cost annealing method still suffers from KL vanishing, because we use the Transformer~\cite{vaswani2017attention} as the decoder, which can predict the current data based on previous ground-truth data. Fig.~\ref{fig:KL-vanish} (c) illustrates the tuning result of $\beta(t)$ by ControlVAE compared with other methods. We can discover that our $\beta(t)$ gradually converges to around a certain value. Note that, here $\beta(t)$ of ControlVAE does not converge to $1$ because we slightly increase the value of KL-divergence (produced by the original VAE) in order to improve the diversity of generated data.

In order to further demonstrate ControlVAE can improve the diversity of generated text, we apply it to dialog-response generation using the Switchboard(SW) dataset. Following~\cite{zhao2017learning}, we adopt a conditional VAE~\cite{zhao2017learning} that generates dialog conditioned on the previous response. We use metric $Dis$-$n$~\cite{xu2018dp} and self-BLEU~\cite{zhu2018texygen} (with 1000 sampled results) to measure the diversity of generated data, and perplexity (PPL)~\cite{jelinek1977perplexity} to measure how well the probability distribution predicts a sample. Table~\ref{tab:dist} illustrates the comparison results for different approaches. We can observe that ControlVAE has more distinct grams and lower self-BLEU than the baselines when the desired KL-divergence is set to $35$ and $25$. In addition, it has lower PPL than the other methods. Thus, we can conclude that ControlVAE can improve the diversity of generated data and generation performance. We also illustrate some examples of generated dialog by ControlVAE in Appendix~\ref{app:exp-dialog}.

%%%--figure for disentanglement
\begin{figure*}[!th]
\centering     %%% not \center
\subfigure[Reconstruction loss]{\includegraphics[width=0.32\textwidth]{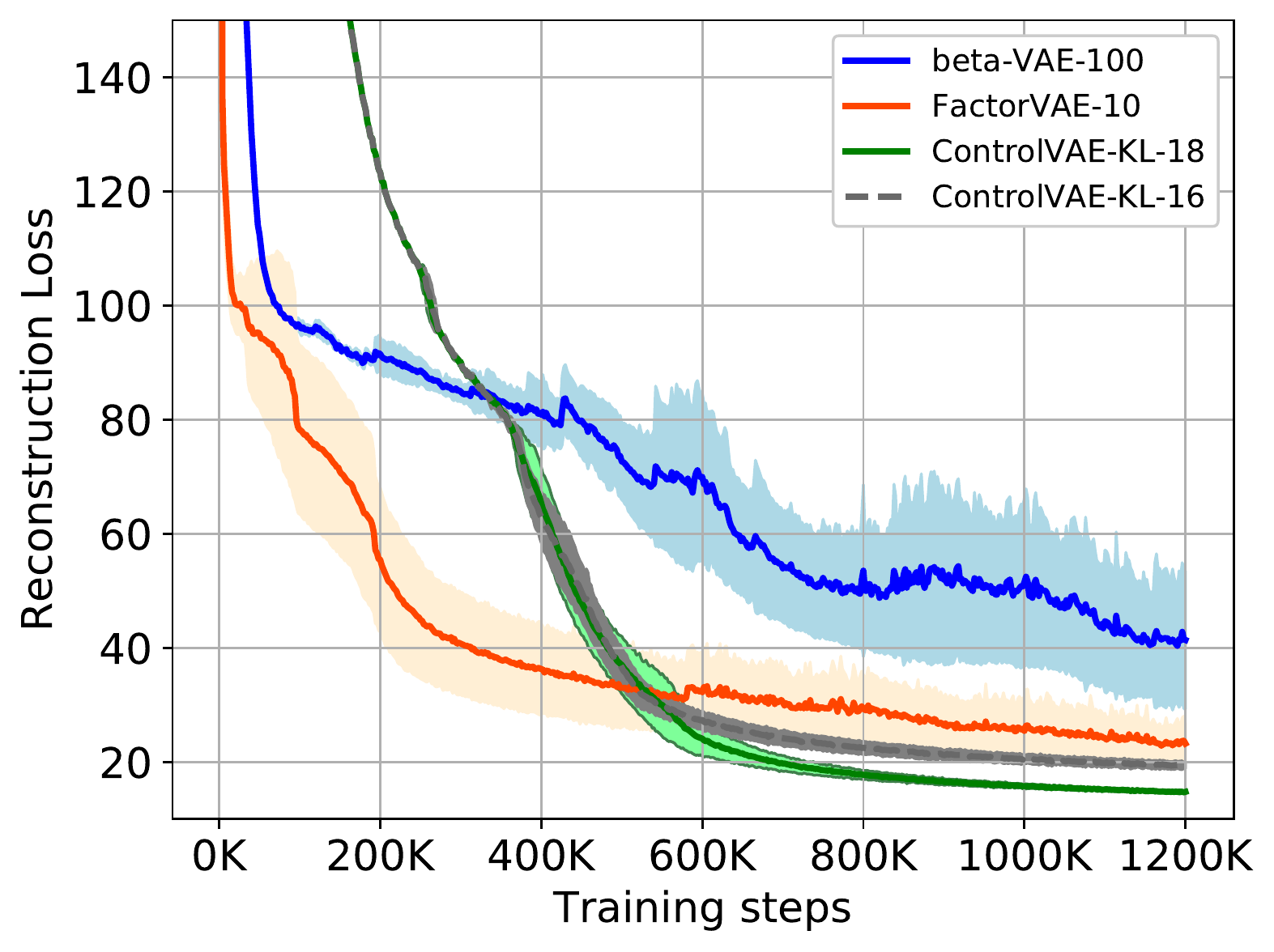}}
\subfigure[$\beta(t)$]{\includegraphics[width=0.32\textwidth]{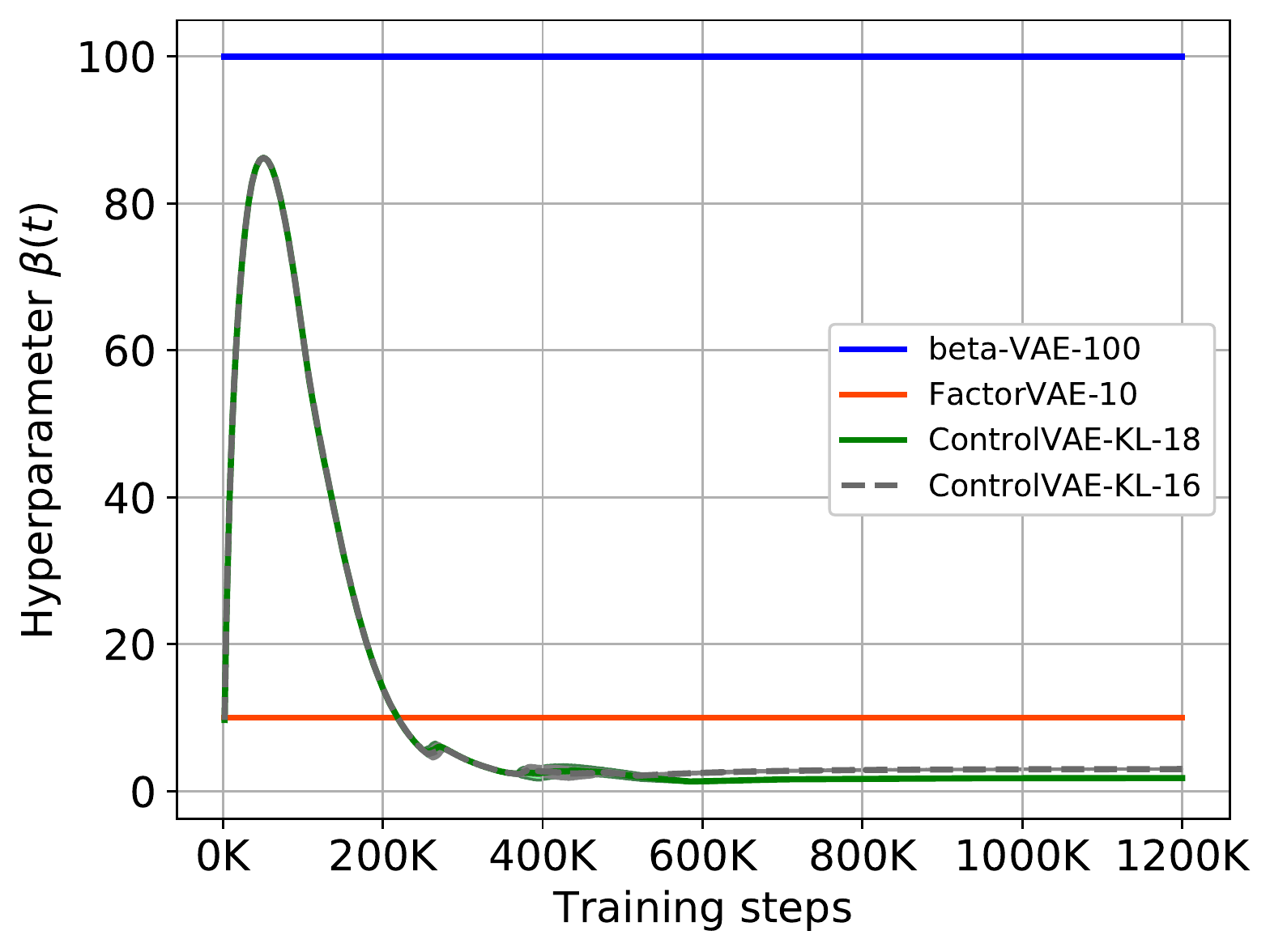}}
\subfigure[Disentangled factors]{\includegraphics[width=0.32\textwidth]{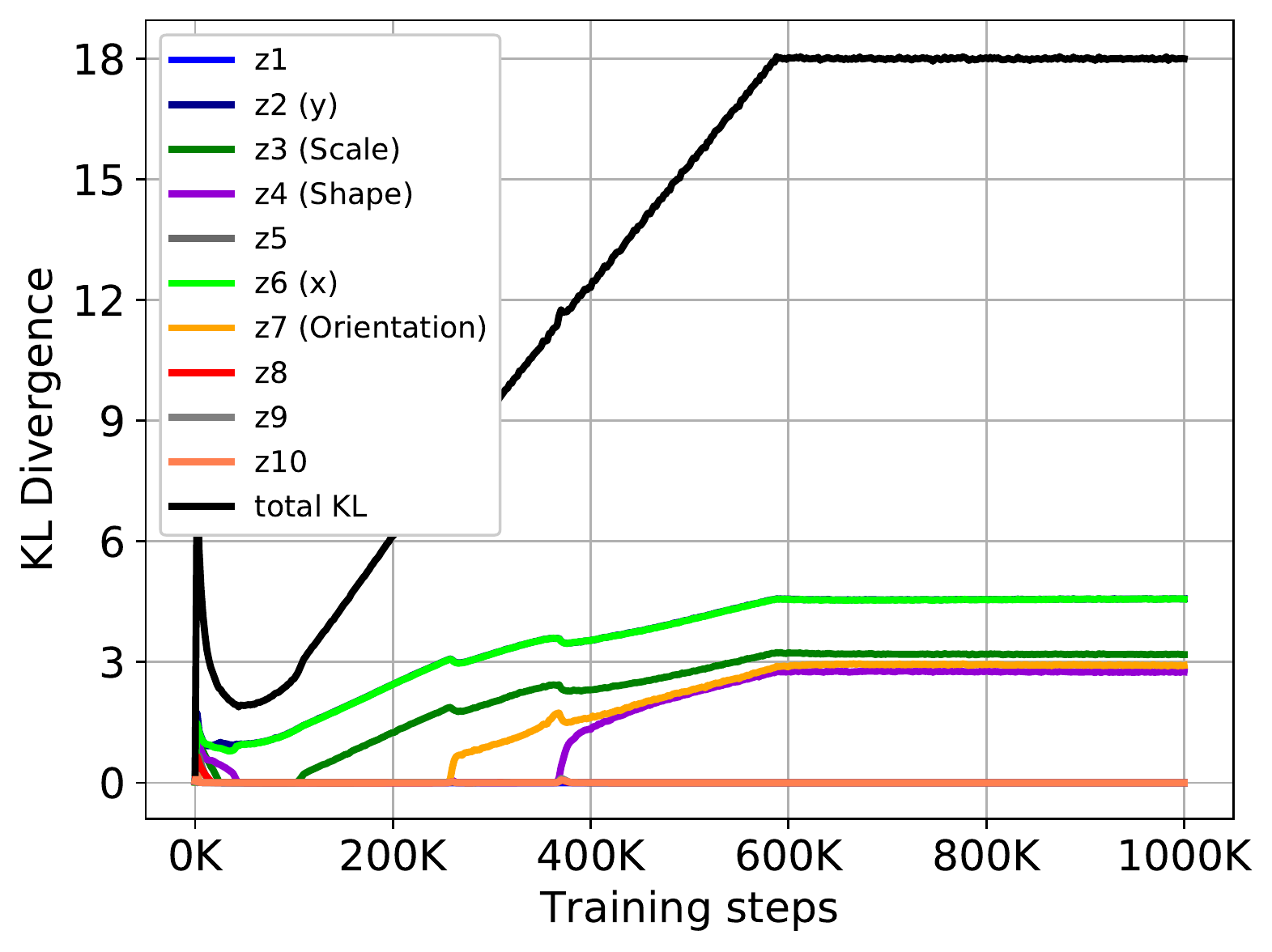}}
\vspace{-0.15in}
\caption{(a) (b) shows the comparison of reconstruction error and $\beta(t)$ using 2D Shapes data over $5$ random seeds. ControlVAE (KL=16, 18) has lower reconstruction errors and variance compared to $\beta$-VAE. (c) shows an example about the disentangled factors in the latent variable as the total KL-divergence increases from $0.5$ to $18$ for ControlVAE (KL=18). Each curve with positive KL-divergence (except black one) represents one disentangled factor by ControlVAE.}\label{fig:disentangled}
\vskip -0.05in
\end{figure*}

%%--table---
\begin{table*}[!ht]
\caption{Performance comparison of different methods using disentanglement metric, MIG score, averaged over 5 random seeds. The higher is better. ControlVAE (KL=16) has a comparable MIG score but lower variance than the FactorVAE with the default parameters.}
\label{tab:mig}
\vskip -0.05in
\begin{center}
\begin{small}
\begin{tabular}{lllll}
\toprule
Metric & ControlVAE (KL=16) & ControlVAE (KL=18) & $\beta$-VAE ($\beta=100$) & FactorVAE ($\gamma=10$) \\
\midrule
MIG & \textbf{ 0.5628 $\pm$ 0.0222}& 0.5432 $\pm$ 0.0281 & 0.5138 $\pm$ 0.0371 & 0.5625 $\pm$ 0.0443 \\
\bottomrule
\end{tabular}
\end{small}
\end{center}
\vskip -0.1in
\end{table*}

%% plot qualitative results %%
\begin{figure*}[!ht]
\begin{center}
\centerline{\includegraphics[width=\textwidth]{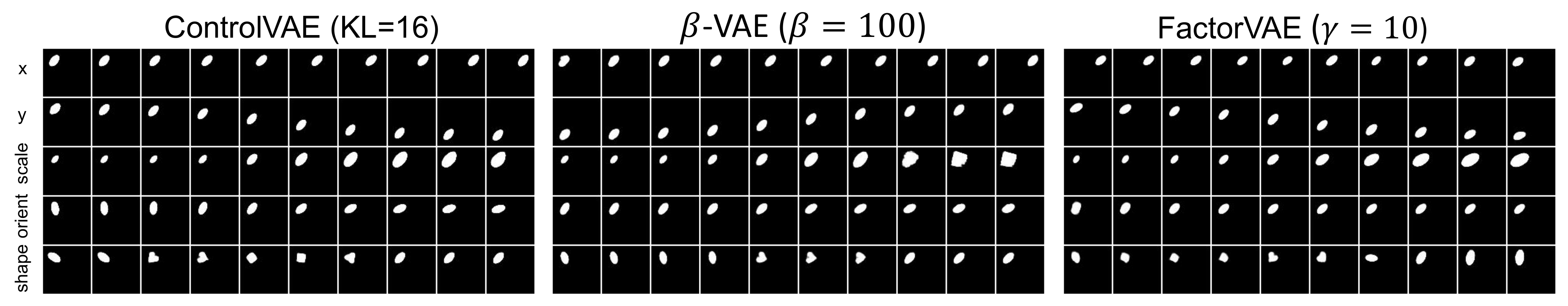}}
%\vspace{-0.05in}
\caption{Rows: latent traversals ordered by the value of KL-divergence with the prior in a descending order. Following work~\cite{burgess2018understanding}, we initialize the latent representation from a seed image, and then traverse a single latent dimension in a range of $[-3,3]$, while keeping the remaining latent dimensions fixed. ControlVAE can disentangle all the five generative factors for 2D Shapes data, while $\beta$-VAE entangles the scale and shape (in 3rd row) and FactorVAE does not disentangle orientation (in 4th row) very well.}
\label{fig:sprites_image}
\end{center}
\vskip -0.15in
\end{figure*}
%%%%%end of figure%%

\subsection{Evaluation on Disentangled Representations}
We then evaluate the performance of ControlVAE on the learning of disentangled representations using \textit{2D Shapes} data. We compare it with two baselines: FactorVAE~\cite{kim2018disentangling} and $\beta$-VAE~\cite{burgess2018understanding}. 

Fig.~\ref{fig:disentangled} (a) and (b) shows the comparison of reconstruction error and the hyperparameter $\beta(t)$ (using $5$ random seeds) for different models. We can observe from Fig.~\ref{fig:disentangled} (a) that ControlVAE (KL=16,18) has lower reconstruction error and variance than the baselines. This is because our ControlVAE automatically adjusts the hyperparameter, $\beta(t)$, to stabilize the KL-divergence, while the other two methods keep the hyperparameter unchanged during model training. Specifically, for ControlVAE (KL=18), the hyperparameter $\beta(t)$ is high in the beginning in order to obtain good disentangling, and then it gradually drops to around $1.8$ as the training converges, as shown in Fig.~\ref{fig:disentangled}(b). In contrast, $\beta$-VAE ($\beta=100$) has a large and fixed weight on the KL-divergence so that its optimization algorithm tends to optimize the KL-divergence term, leading to a large reconstruction error. In addition, Fig.~\ref{fig:disentangled}(c) illustrates an example of KL-divergence per factor in the latent code as training progresses and the total information capacity (KL-divergence) increases from $0.5$ until to $18$. We can see that ControlVAE disentangles all the five generative factors, starting from positional latents ($x$ and $y$) to scale, followed by orientation and then shape.

To further demonstrate ControlVAE can achieve a better disentangling, we use a disentanglement metric, mutual information gap (MIG)~\cite{chen2018isolating}, to compare their performance, as shown in Table~\ref{tab:mig}. It can be observed that ControlVAE (KL=16) has a comparable MIG but lower variance than FactorVAE. Here it is worth noting that FactorVAE adds a Total Correlation (TC) term in the objective while our method does not. Since there does not exist a robust metric to fully measure disentanglement, we also show the qualitative results of different models in Fig.~\ref{fig:sprites_image}. We can observe that ControlVAE can discover all the five generative factors: positional latent ($x$ and $y$), scale, orientation and shape. However, $\beta$-VAE ($\beta=100$) disentangles four generative factors except for entangling the scale and shape together (in the third row), while FactorVAE ($\gamma=10$) does not disentangle the orientation factor very well in the fourth row in Fig.~\ref{fig:sprites_image}. Thus, ControlVAE achieves a better reconstruction quality and disentangling than the baselines.

%%---image generation---
\subsection{Evaluation on Image Generation}
\begin{figure}[htb]
\centering     %%% not \center
\subfigure[Reconstruction loss]{\includegraphics[width=0.49\columnwidth]{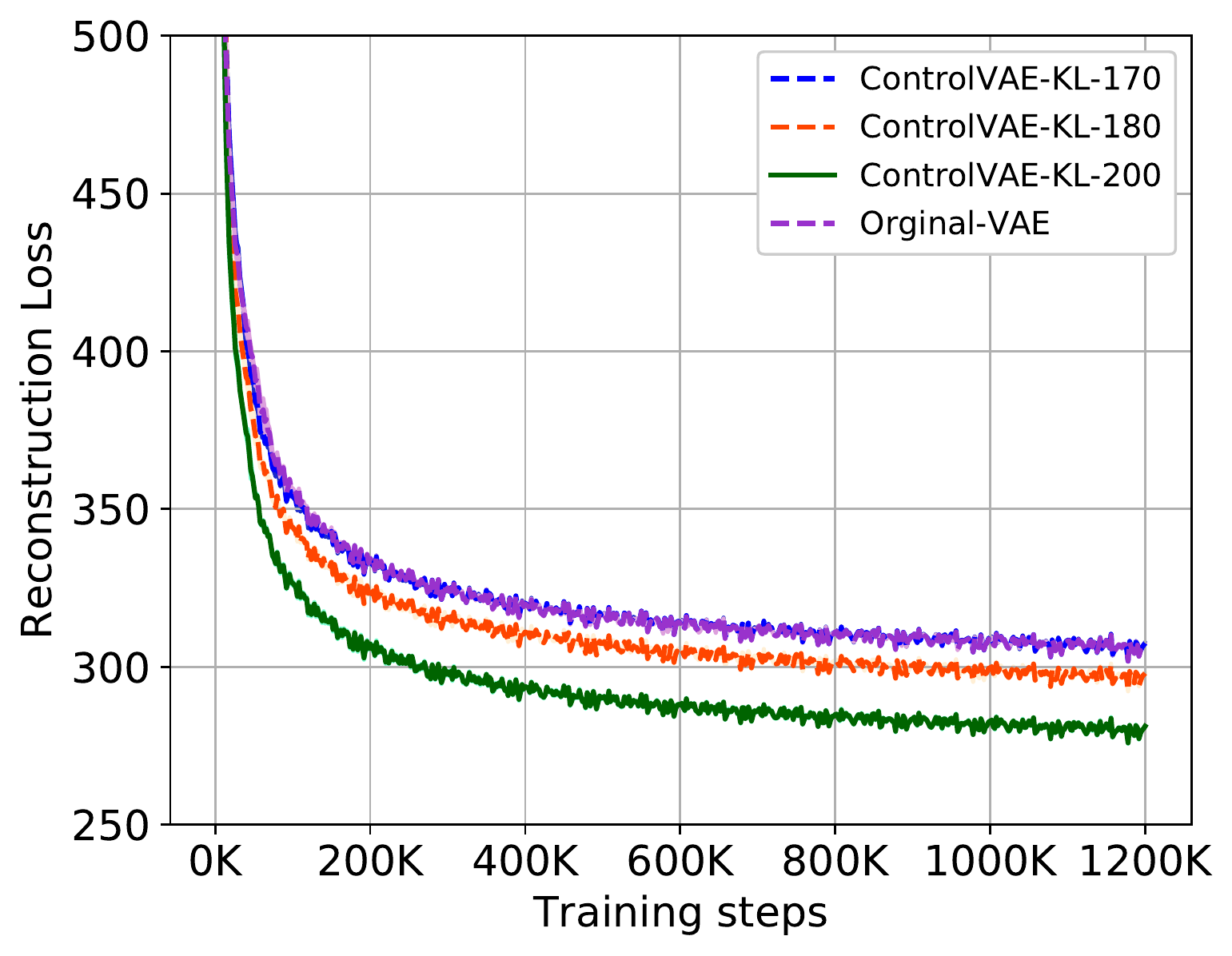}}
\subfigure[KL-divergence]{\includegraphics[width=0.49\columnwidth]{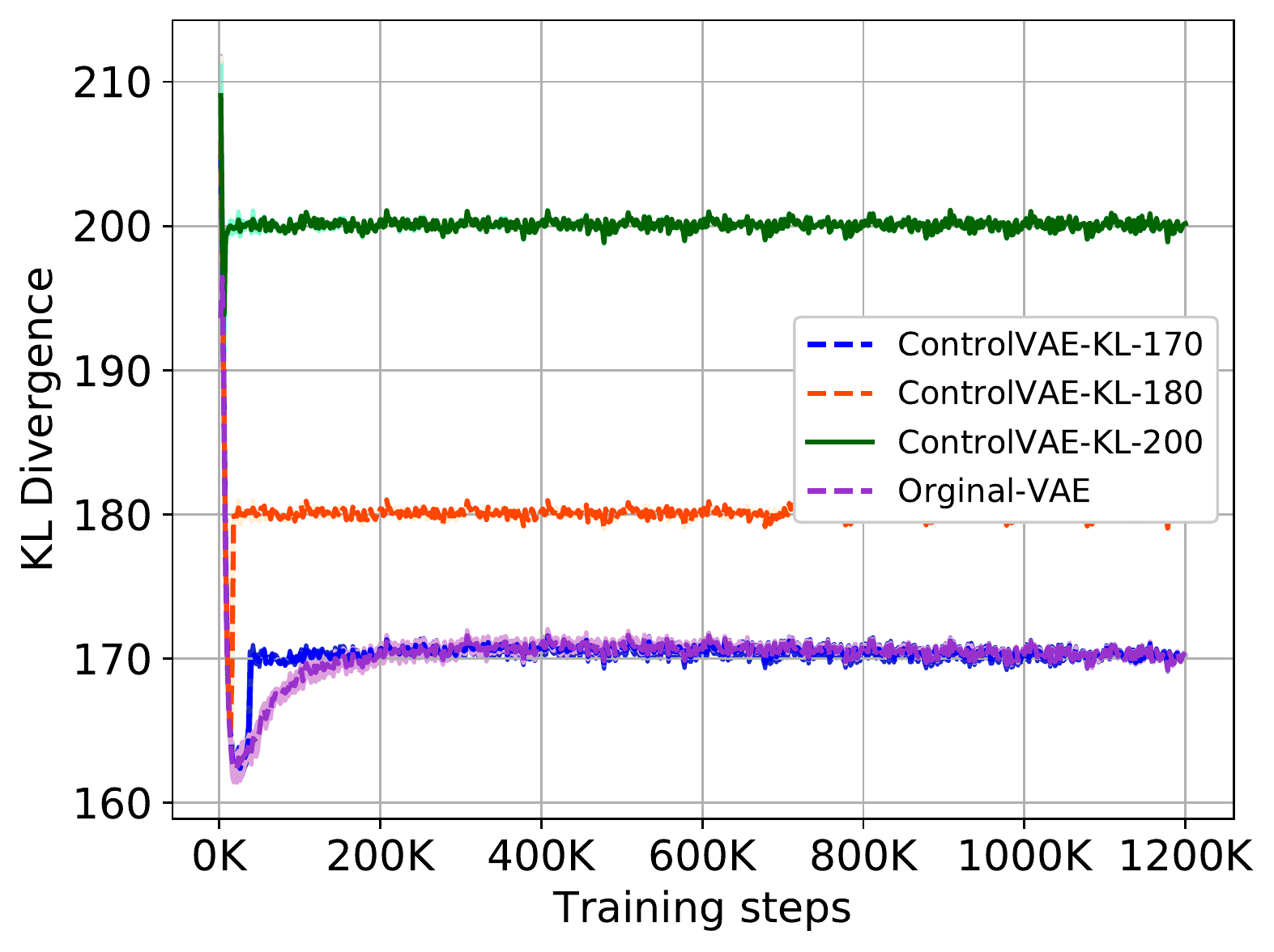}}
%\vspace{-0.05in}
\caption{Performance comparison for different methods on the CelebA data. }\label{fig:CelebA-compare}
%\vskip -0.01in
\end{figure}

Finally, we compare the reconstruction quality on image generation task for ControlVAE and the original VAE. Fig.~\ref{fig:CelebA-compare} shows the comparison of reconstruction error and KL-divergence under different desired values of KL-divergence averaged over 3 random seeds. We can see from Fig.~\ref{fig:CelebA-compare}(a) that ControlVAE-KL-200 (KL=200) has the lowest reconstruction error among them. This is because there exists a trade-off between the reconstruction accuracy and KL-divergence in the VAE objective. When the value of KL-divergence rises, which means the weight of KL term decreases, the model tends to optimize the reconstruction term. In addition, as we set the desired KL-divergence to $170$ (same as the basic VAE in Fig.~\ref{fig:CelebA-compare}(b)), ControlVAE has the same reconstruction error as the original VAE. At that point, ControlVAE becomes the original VAE as $\beta(t)$ finally converges to $1$, as shown in Fig.~\ref{fig:beta-image} in Appendix~\ref{app:weight-image}.

We further adopt two commonly used metrics for image generation, FID~\cite{lucic2018gans} and SSIM~\cite{chen2018attention}, to evaluate the performance of ControlVAE in Table~\ref{tab:metrics}. It can be observed that ControlVAE-KL-200 outperforms the other methods in terms of FID and SSIM. Therefore, ControlVAE can improve the reconstruction quality for image generation task via  slightly increasing the value of KL-divergence. We also show some examples to verify ControlVAE has a better reconstruction quality than the basic VAE in Appendix~\ref{app:image_exp}.

%%--celebA image--

%%--table---
\begin{table}[ht]
\caption{Performance comparison for different methods on CelebA data over $3$ random seeds. FID: lower is better. SSIM: higher is better.}
\label{tab:metrics}
% \vskip 0.05in
\begin{center}
\begin{small}
% \begin{sc}
\begin{tabular}{lllll}
\toprule
Methods/metric & FID & SSIM  \\
\midrule
ControlVAE-KL-200   & \textbf{55.16} $\pm$ 0.187 &  \textbf{0.687} $\pm$ 0.0002  \\
ControlVAE-KL-180 & 57.57 $\pm$ 0.236 & 0.679 $\pm$ 0.0003  \\
ControlVAE-KL-170  & 58.75 $\pm$ 0.286 & 0.675 $\pm$ 0.0001   \\
Original VAE    & 58.71 $\pm$ 0.207 &0.675 $\pm$ 0.0001 \\
\bottomrule
\end{tabular}
% \end{sc}
\end{small}
\end{center}
\vskip -0.1in
\end{table}

%% file: relatedwork.tex
\section{Related Work}
\label{sec:relatedwork}
% We review the related work about VAE and its variants in various applications.
We review the related work about VAE and its variants in various applications, and then point out the difference between our method and the prior work.

There are many work involving a trade-off between reconstruction and KL-divergence for VAEs applications. For disentangled representation learning, researchers proposed $\beta$-VAE ($\beta > 1$)~\cite{higgins2017beta,burgess2018understanding} that assigns a large and fixed hyperparameter, $\beta$, to put more emphasis on the KL divergence to encourage disentangled latent representations. It, however, sacrifice the reconstruction quality in order to obtain better disentangling. Then some follow-up work~\cite{chen2018isolating,kim2018disentangling} further factorize the KL-divergence term to improve the reconstruction quality. However, these methods still assign a fixed and large hyperparameter to the decomposed terms in the objective, resulting in high reconstruction error. In contrast, ControlVAE dynamically tunes $\beta$ during optimization to achieve good disentangling \textit{and} better reconstruction quality. More importantly, our PI control method can be used as a plug-in replacement of existing methods.

In order to improve the sample generation quality of VAEs~\cite{dai2019diagnosing,zhao2019infovae,xiao2019generative,ghosh2019variational,alemi2017fixing,zhao2019infovae}, some researchers tried to reduce the weight of KL-divergence to make the decoder produce sharper outputs. Though they can obtain impressive sample quality, they suffer severely from the trade-off in the way that the latent distribution is far away from the prior. Recent studies adopted a constrained optimization for reconstruction error~\cite{rezende2018taming,klushyn2019learning} to achieve the trade-off between reconstruction error and KL-divergence. However, they may suffer from posterior collapse if the inference network fails to cover the latent space while our can totally avert posterior collapse. In addition, different from their work, we try to optimize KL-divergence (information bottleneck) as a constraint. Our method and theirs complement each other for different applications.

In language modeling, VAE often suffers from KL vanishing, due to a powerful decoder, such as Transformer~\cite{vaswani2017attention} and LSTM. To remedy this issue, one popular way is to add a hyperparameter $\beta$ on the KL term~\cite{bowman2015generating,liu2019cyclical}, and then gradually increases it from $0$ until $1$. However, the existing methods~\cite{yang2017improved,bowman2015generating,liu2019cyclical}, such as KL cost annealing and cyclical annealing, cannot totally solve KL vanishing or explicitly control the value of KL-divergence since they blindly change $\beta$ without observing the actual KL-divergence during model training. Conversely, our approach can avert KL vanishing and stabilize the KL-divergence to a desired value.

%% file: Conclusion.tex
\section{Conclusion and Future Work}
\label{sec:conclusion}
In this paper, we proposed a general controllable VAE framework, ControlVAE, that combines automatic control with the basic VAE framework to improve the performance of the VAE models. We designed a new non-linear PI controller to control the value of KL divergence during model training. Then we evaluated ControlVAE on three different tasks. The results show that ControlVAE attains better performance; it can achieve a higher reconstruction quality and good disentanglement. It can totally avert KL vanishing and improve the diversity of generated data in language modeling. In addition, it improves the reconstruction quality on the task of image generation. For future work, we plan to apply our method to other research topics, such as topic modeling and semi-supervised applications.

%% file: acknowledge.tex
\section*{Acknowledgments}
Research reported in this paper was sponsored in part by DARPA award W911NF-17-C-0099, DTRA award HDTRA1-18-1-0026, and the Army Research Laboratory under Cooperative Agreements W911NF-09-2-0053 and W911NF-17-2-0196.

%% file: appendix.tex
\appendix

\section{Model Configurations and hyperparameter settings}
\label{sec:configure}
We summarize the detailed model configurations and hyperparameter settings for ControlVAE in the following three applications: language modeling, disentanglement representation learning and image generation.

\subsection{Experimental Details for Language Modeling}
For text generation on PTB data, we build the ControlVAE model on the basic VAE model, as in~\cite{bowman2015generating}. We use one-layer LSTM as the encoder and a three-layer Transformer with eight heads as the decoder and a Multi-Layer Perceptron (MLP) to learn the latent variable~$\mathbf{z}$. The maximum sequence length for LSTM and Transformer is set to $100$, respectively. And the size of latent variable is set to $64$. Then we set the dimension of word embedding to $256$ and the batch size to $32$. In addition, the dropout is $0.2$ for LSTM and Transformer. Adam optimization with the learning rate $0.001$ is used during training. Following the tuning guidelines above, we set the coefficients $K_p$ and $K_i$ of P term and I term to $0.01$ and $0.0001$, respectively. Finally, We adopt the source code on Texar platform to implement experiments~\cite{hu2019texar}. 

For dialog-response generation, we follow the model architecture and hyperparameters of the basic conditional VAE in~\cite{zhao2017learning}. We use one-layer Bi-directional GRU as the encoder and one-layer GRU as the decoder and two fully-connected layers to learn the latent variable. In the experiment, the size of both latent variable and word embeddings is set to $200$. The maximum length of input/output sequence for GRU is set to $40$ with batch size $30$. In addition, Adam with initial learning rate $0.001$ is used. In addition, we set the same $K_p$ and $K_i$ of PI algorithm as text generation above.
The model architectures of ControlVAE for these two NLP tasks are illustrated in Table~\ref{tab:ptb_model},~\ref{tab:sw_model}.

%%%%--PTB model--%%%%
\begin{table}[htb]
\caption{Encoder and decoder architecture for text generation on PTB data.}
\label{tab:ptb_model}
\begin{center}
\begin{small}
\begin{tabular}{|l|l|}
\toprule
Encoder & Decoder \\
\midrule
Input $n$ words $\times 256$ & Input $\in \mathbb{R}^{64}$, $n \times 256$ \\
\midrule
1-layer LSTM & FC $64 \times 256$ \\
\midrule
FC $64 \times 2$ & 3-layer Transformer 8 heads \\
\bottomrule
\end{tabular}
\end{small}
\end{center}
% \vskip -0.05in
\end{table}

%%%---SW dataset--%%%
\begin{table}[htb]
\caption{Encoder and decoder architecture for dialog generation on Switchboard (SW) data.}
\label{tab:sw_model}
\begin{center}
\begin{small}
\begin{tabular}{|l|l|}
\toprule
Encoder & Decoder \\
\midrule
\textnormal{Input $n$ words $\times 200$} & \textnormal{Input $\in \mathbb{R}^{200}$}  \\
\midrule
\textnormal{1-layer bi-GRU} & \textnormal{FC $200 \times 400$} \\
\midrule
\textnormal{FC $200 \times 2$} & \textnormal{1-layer GRU} \\
\midrule
\textnormal{FC $200 \times 2$} & \\
\bottomrule
\end{tabular}
\end{small}
\end{center}
\vskip -0.1in
\end{table}

\subsection{Experimental Details for Disentangling}
Following the same model architecture of $\beta$-VAE~\cite{higgins2017beta}, we adopt a convolutional layer and deconvolutional layer for our experiments. We use Adam optimizer with $\beta_1=0.90$, $\beta_2=0.99$ and a learning rate tuned from $10^{-4}$. We set $K_p$ and $K_i$ for PI algorithm to $0.01$ and $0.001$, respectively. For the step function, we set the step, $\alpha$, to $0.15$ per $K=5000$ training steps as the information capacity (desired KL- divergence) increases from $0.5$ until $18$ for 2D Shape data. ControlVAE uses the same encoder and decoder architecture as $\beta$-VAE except for plugging in PI control algorithm, illustrated in Table~\ref{tab:2Dshape_model}.

%%%%%2D shapes%%%
\begin{table}[htb]
\caption{Encoder and decoder architecture for disentangled representation learning on 2D Shapes data.}
\label{tab:2Dshape_model}
\begin{center}
\begin{scriptsize}
\begin{tabular}{|l|l|}
\toprule
Encoder & Decoder \\
\midrule
Input $64\times64$ binary image  & Input $\in \mathbb{R}^{10}$ \\
\midrule
$4\times4$ conv. $32$ ReLU. stride 2 &  FC. 256 ReLU.\\
\midrule
$4\times4$ conv. $32$ ReLU. stride 2 & $4\times4$ upconv. $256$ ReLU. \\
\midrule
$4\times4$ conv. $64$ ReLU. stride 2 &  $4\times4$ upconv. $64$ ReLU. stride 2.\\
\midrule
$4\times4$ conv. $64$ ReLU. stride 2 & $4\times4$ upconv. $64$ ReLU. stride 2 \\
\midrule
$4\times4$ conv. $256$ ReLU. &  $4\times4$ upconv. $32$ ReLU. stride 2 \\
\midrule
FC $256$. FC. $2 \times 10$ &  $4\times4$ upconv. $32$ ReLU. stride 2 \\
\bottomrule
\end{tabular}
\end{scriptsize}
\end{center}
% \vskip -0.1in
\end{table}

\subsection{Experimental Details for Image Generation}
Similar to the architecture of $\beta$-VAE, we use a convolutional layer with batch normalization as the encoder and a deconvolutional layer with batch normalization for our experiments. We use Adam optimizer with $\beta_1=0.90$, $\beta_2=0.99$ and a learning rate $10^{-4}$ for CelebA data. The size of latent variable is set to $500$, because we find it has a better reconstruction quality than $200$ and $400$. In addition, we set the desired value of KL-divergence to $170$ (same as the original VAE), $180$, and $200$. For PI control algorithm, we set $K_p$ and $K_i$ to $0.01$ and $0.0001$, respectively. We also use the same encoder and decoder architecture as $\beta$-VAE above except that we add the batch normalization to improve the stability of model training, as shown in Table~\ref{tab:celeba_model}. 

%%%%%%CelebA %%
\begin{table}[htb]
\caption{Encoder and decoder architecture for image generation on CelebA data.}
\label{tab:celeba_model}
\begin{center}
\begin{scriptsize}
\begin{tabular}{|l|l|}
\toprule
Encoder & Decoder \\
\midrule
Input $128\times 128 \times 3$ RGB image  & Input $\in \mathbb{R}^{500}$ \\
\midrule
$4\times4$ conv. $32$ ReLU. stride 2 &  FC. 256 ReLU.\\
\midrule
$4\times4$ conv. $32$ ReLU. stride 2 & $4\times4$ upconv. $256$ ReLU. stride 2 \\
\midrule
$4\times4$ conv. $64$ ReLU. stride 2 &  $4\times4$ upconv. $64$ ReLU. stride 2.\\
\midrule
$4\times4$ conv. $64$ ReLU. stride 2 & $4\times4$ upconv. $64$ ReLU. stride 2 \\
\midrule
$4\times4$ conv. $256$ ReLU. stride 2 &  $4\times4$ upconv. $32$ ReLU. stride 2 \\
\midrule
FC $4096$. FC.$2 \times 500$ &  $4\times4$ upconv. $32$ ReLU. stride 2 \\
\bottomrule
\end{tabular}
\end{scriptsize}
\end{center}
\vskip -0.1in
\end{table}

\section{Examples of Generated Dialog by ControlVAE}
\label{app:exp-dialog}
In this section, we show an example to compare the diversity and relevance of generated dialog by different methods, as illustrated in Table~\ref{tab:dialog_exp}. Alice begins with the open-ended conversation on choosing a college. Our model tries to predict the response from Bob. The ground truth response is ``um - hum''. We can observe from Table~\ref{tab:dialog_exp} that ControlVAE (KL=25, 35) can generate diverse and relevant response compared with the ground truth. In addition, while cyclical annealing can generate diverse text, some of them are not very relevant to the ground-truth response.

\begin{table}[htb]
\caption{Examples of generated dialog for different methods. Our model tries to predict the response from Bob. The response generated by ControlVAE (KL=25,35) are relevant and diverse compared with the ground truth. However, some of reponse generated by cost annealing and cyclical annealing are not very relevant to the ground-truth data}
\vskip -0.1in
\label{tab:dialog_exp}
\begin{center}
\begin{small}
\begin{tabular}{p{0.225\textwidth}|p{0.225\textwidth}}
\toprule
\multicolumn{2}{p{0.44\textwidth}}{\textbf{Context}: (Alice) and a lot of the students in that home town sometimes $\langle$ unk $\rangle$ the idea of staying and going to school across the street so to speak} \\
% \midrule
\multicolumn{2}{l}{\textbf{Topic}: Choosing a college \quad \textbf{Target}: (Bob) um - hum} \\
\midrule
\textbf{ControlVAE-KL-25} &  \textbf{ControlVAE-KL-35}  \\
 \midrule
yeah &  uh - huh   \\
  \midrule
um - hum &  yeah \\
  \midrule
oh that's right um - hum & oh yeah oh absolutely \\
  \midrule
yes & right\\
  \midrule
right & um - hum\\
\midrule
\textbf{Cost annealing (KL=17)} &  \textbf{Cyclical anneal (KL=21.5)}  \\
\midrule
oh yeah &  yeah that's true do you do you do it    \\
 \midrule
uh - huh &  yeah\\
\midrule
right & um - hum\\
\midrule
uh - huh and i think we have to be together & yeah that's a good idea\\
\midrule
oh well that's neat yeah well & yeah i see it too,it's a neat place\\
\bottomrule
\end{tabular}
\end{small}
\end{center}
% \vskip -0.1in
\end{table}

\section{$\beta(t)$ of ControlVAE for Image Generation on CelebA data}
\label{app:weight-image}
\begin{figure}[!ht]
%\vskip 0.05in
\begin{center}
\centerline{\includegraphics[width=0.9\columnwidth]{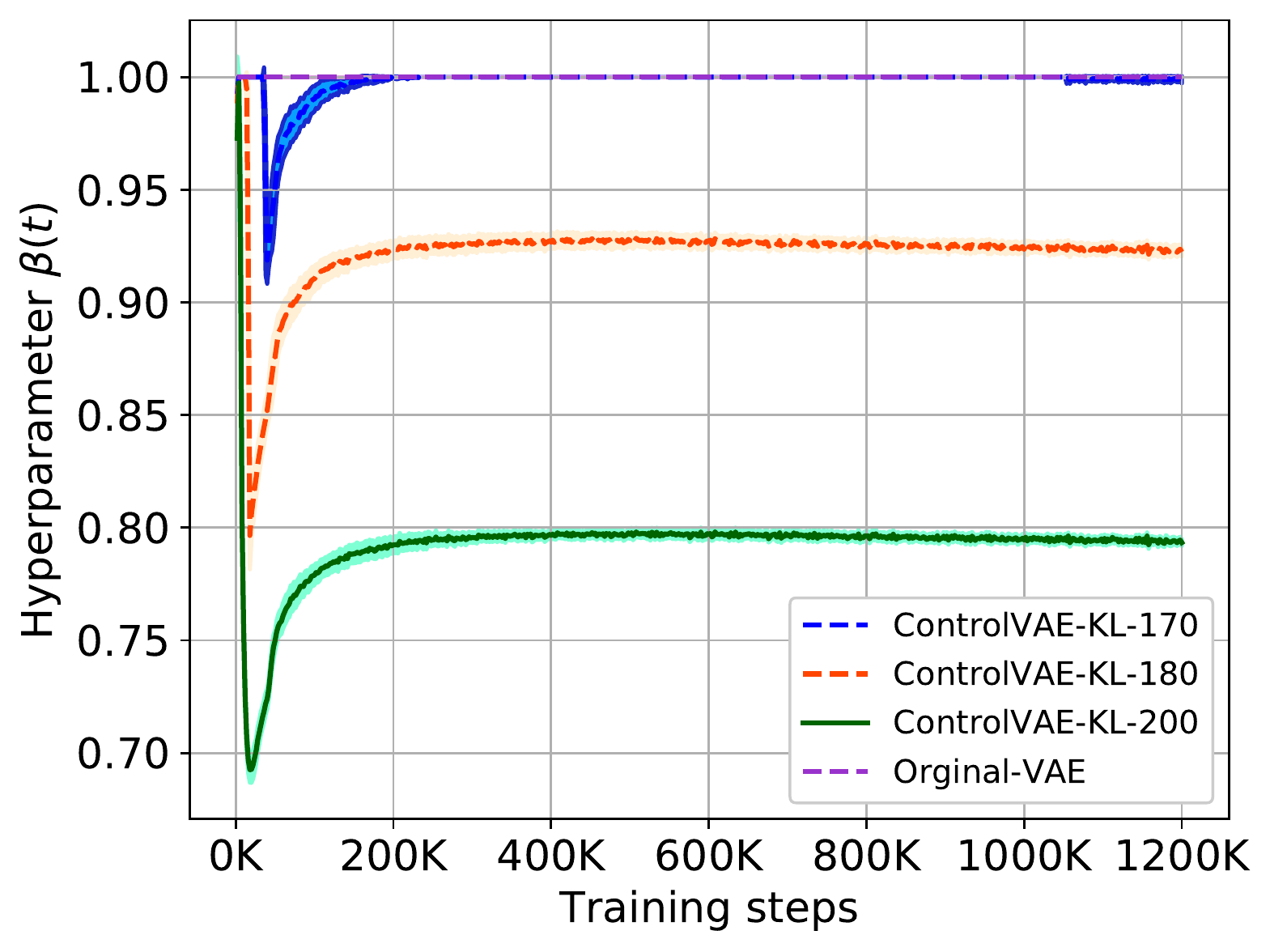}}
\vspace{-0.15in}
\caption{Hyperparameter $\beta(t)$ of ControlVAE for image generation on CelebA data for 3 random seeds. If we set the desired value of KL-divergence to $170$, the hyperparameter, $\beta(t)$, gradually approaches $1$. It means the ControlVAE becomes the original VAE.}
\label{fig:beta-image}
\end{center}
\vskip -0.15in
\end{figure}

Fig.~\ref{fig:beta-image} illustrates the comparison of $\beta(t)$ for different methods during model training. We can observe that $\beta(t)$ finally converges to $1$ when the desired value of KL-divergence is set to $170$, same as the original VAE. At this point, ControlVAE becomes the original VAE. Thus, ControlVAE can be customized by users based on different applications.

\section{Examples of Reconstruction Images by VAE and ControlVAE}
\begin{figure*}[!t]
% \vskip -0.3in
\centering     %%% not \center
\subfigure[Ground truth]{\includegraphics[width=0.49\textwidth]{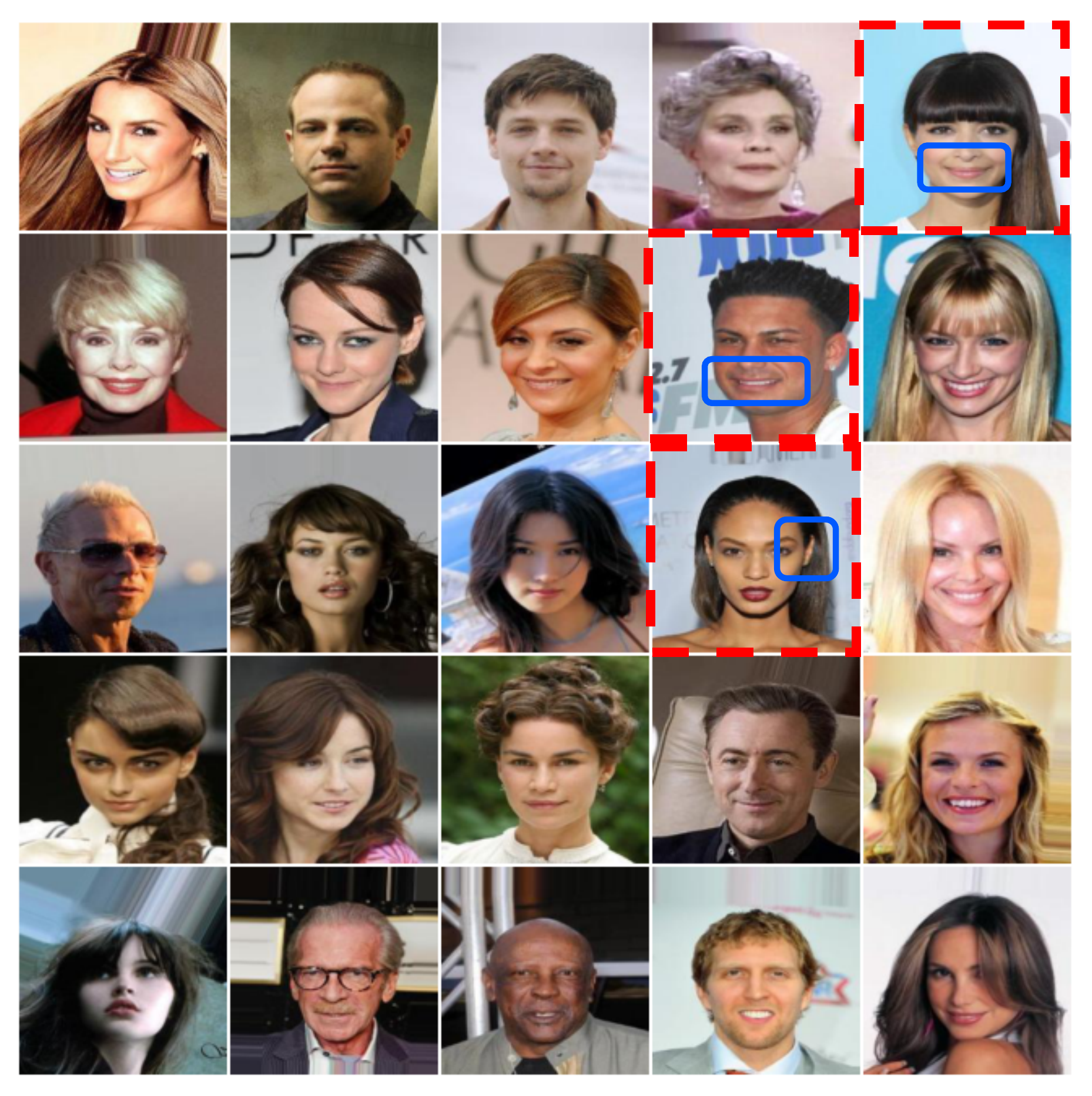}}
\subfigure[Original VAE]{\includegraphics[width=0.49\textwidth]{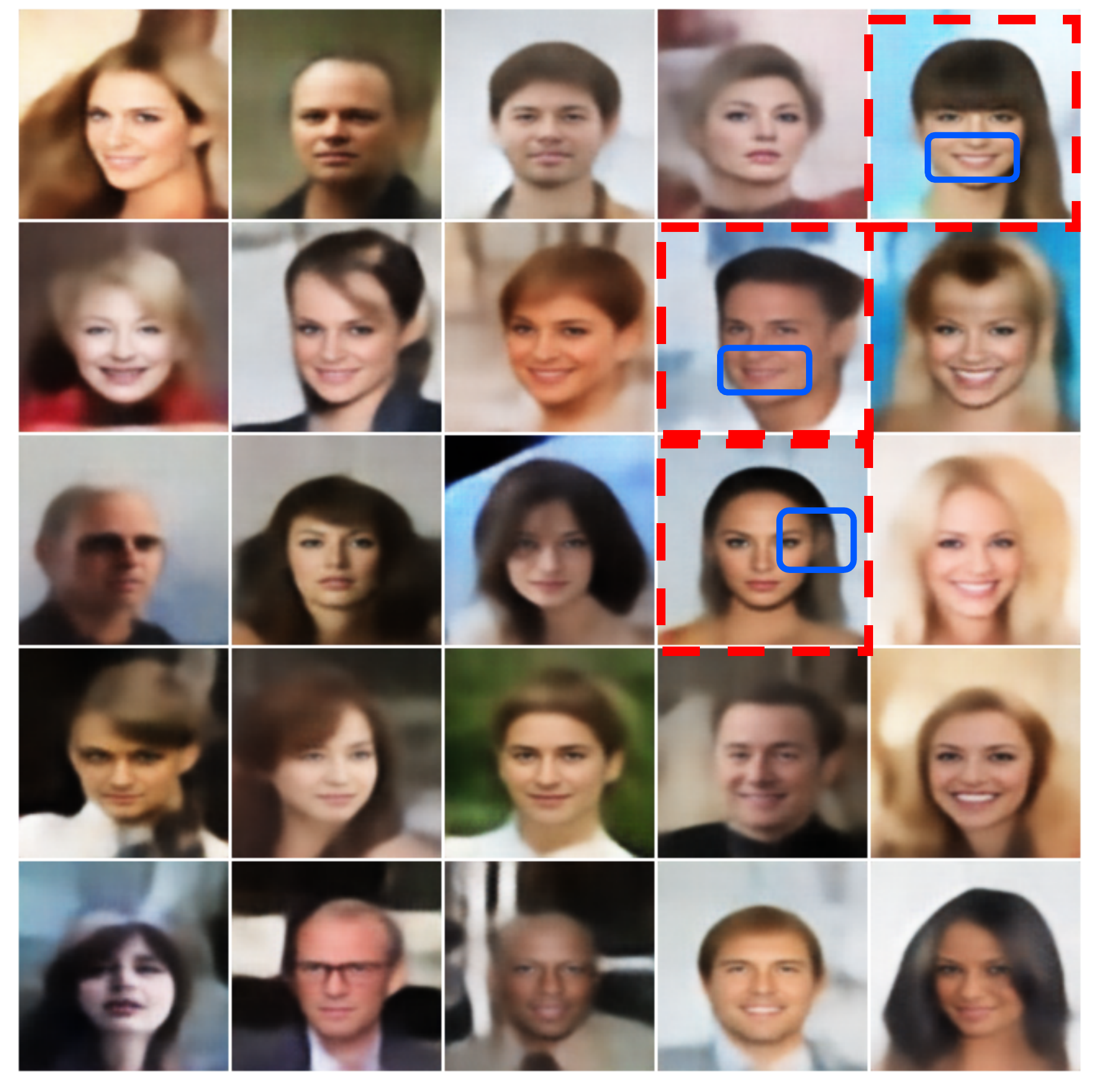}}
\subfigure[ControlVAE-KL-200]{\includegraphics[width=0.49\textwidth]{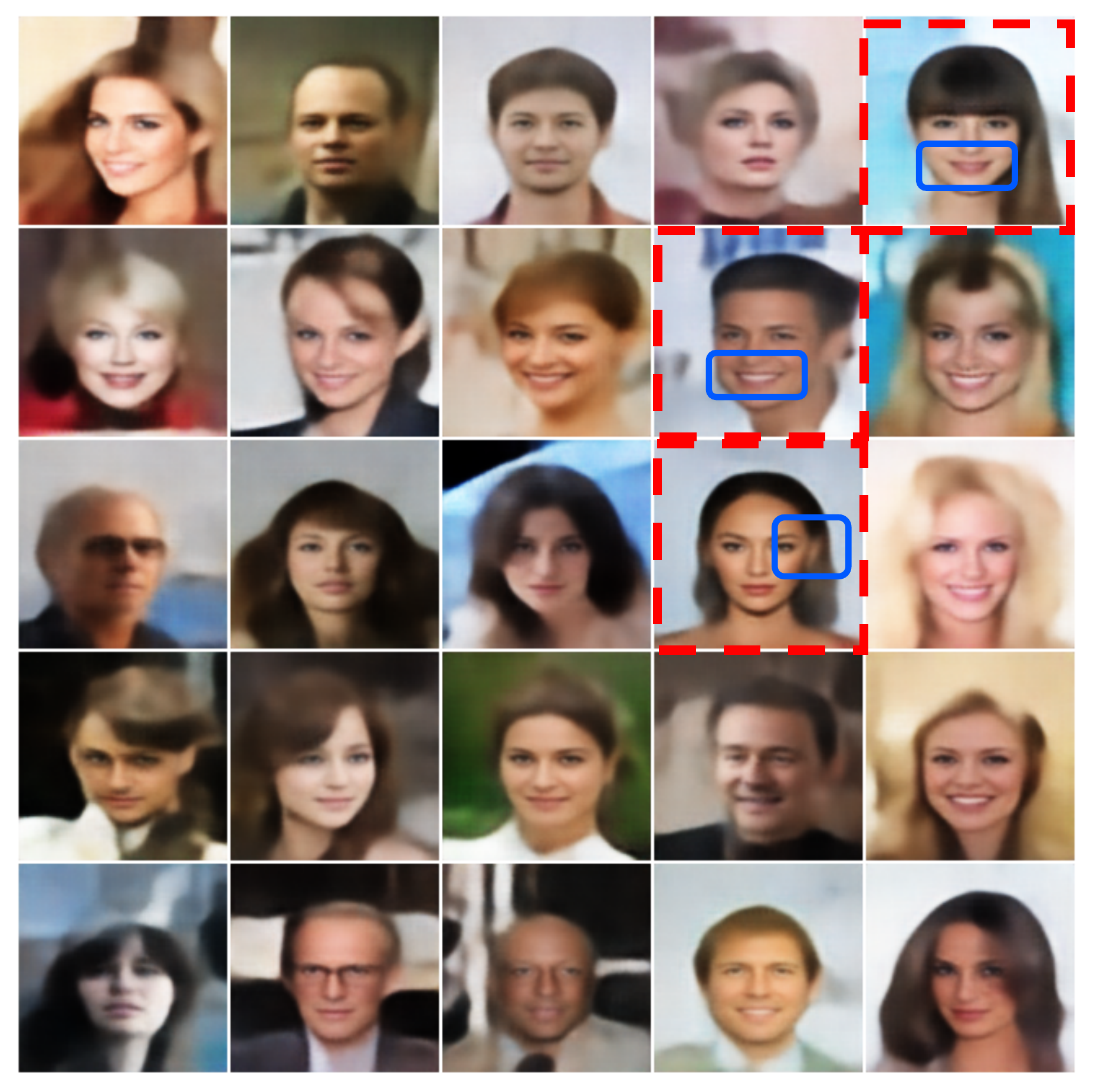}}
\subfigure[ControlVAE-KL-170]{\includegraphics[width=0.49\textwidth]{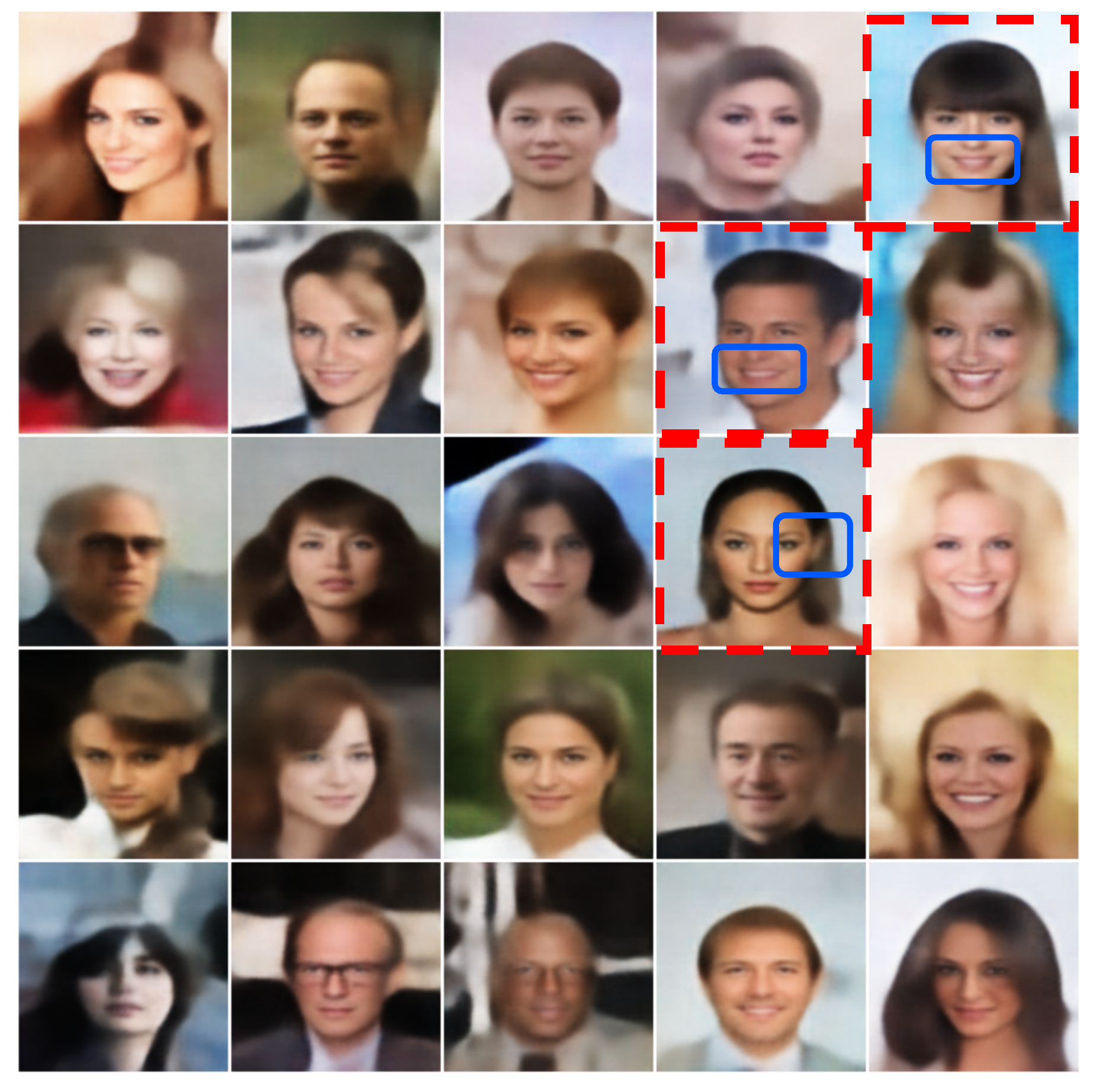}}
% \vskip -0.1in
\caption{Examples of  images recontructed by different methods and ground truth. From the images marked with blue rectangles, we can see that ControlVAE-KL-200 (KL=200) can better reconstruct woman's month opening (first row last column), man's smiling with teeth (second row fourth column), and woman'ear (third row fourth column) than the original VAE based on the ground-truth data in (a).}\label{fig:image_celeba}
% \vskip -0.3in
\end{figure*}

%%%%%paragraph%%%%
We also show some reconstruction images by ControlVAE and the original VAE in Fig.~\ref{fig:image_celeba}. It can be observed that images reconstructed by ControlVAE-KL-200 (KL = $200$) has the best reconstruction quality compared to the original VAE. Take the woman in the first row last column as an example. The woman does not show her teeth in the ground-truth image. However, we can see the woman reconstructed by the original VAE smiles with mouth opening. In contrast, the woman reconstructed by ControlVAE-KL-200 hardly show her teeth when smiling. In addition, we also discover from the other two examples marked with blue rectangles that ControlVAE-KL-200 can better reconstruct the ``smile'' from the man and the ``ear'' from the woman compared to the original VAE. Therefore, we can conclude that our ControlVAE can improve the reconstruction quality via slightly increasing (control) KL-divergence compared to the original VAE. Note that, if we want to improve the generation quality by sampling the latent code, we can reduce the KL divergence, which will be explored in the future.
%
%It should be pointed out that the comparison results are not very obvious because we use a simple VAE model in the experiments. For future work, we are going to adopt advanced VAE models to improve the performance.

\label{app:image_exp}
%%%--figure for celebA
